\documentclass[5p,twocolumn,10pt,times]{elsarticle}
\usepackage{amsmath}
\usepackage{hyperref}
\usepackage{multirow}
\usepackage{caption}
\usepackage{subcaption}
\usepackage{makecell}
\usepackage{siunitx,booktabs,xcolor}
\usepackage[colorinlistoftodos]{todonotes}

\usepackage{fancyhdr,xcolor}
\begin{document}
\baselineskip11pt

\begin{frontmatter}

\title{Real-Time Topology Optimization in 3D via \\ Deep Transfer Learning}

\author[1]{Mohammad Mahdi Behzadi}
\ead{mohammad.behzadi@uconn.edu}

\author[1]{Horea T. Ilie\c{s}\corref{cor1}}
\ead{horea.ilies@uconn.edu}

\address[1]{Computational Design Laboratory (CDL), University of Connecticut, Storrs, CT 06269, USA}

\begin{abstract} 
The published literature on topology optimization has exploded over the last two decades to include methods that use shape and topological derivatives or evolutionary algorithms formulated on various geometric representations and parametrizations. One of the key challenges of all these methods is the massive computational cost associated with 3D topology optimization problems. 

We introduce a transfer learning method based on a convolutional neural network that (1) can handle high-resolution 3D design domains of various shapes and topologies; (2) supports real-time design space explorations as the domain and boundary conditions change; (3) requires a much smaller set of high-resolution examples for the improvement of learning in a new task compared to traditional deep learning networks; (4) is multiple orders of magnitude more efficient than the established gradient-based methods, such as SIMP. We provide numerous 2D and 3D examples to showcase the effectiveness and accuracy of our proposed approach, including for design domains that are \textit{unseen} to our source network, as well as the generalization capabilities of the transfer learning-based approach. Our experiments achieved an average binary accuracy around 95\% at real-time prediction rates. These properties, in turn, suggest that the proposed transfer-learning method may serve as the first practical underlying framework for real-time 3D design exploration based on topology optimization. 

\end{abstract}

\begin{keyword} Topology optimization, Transfer learning, Real-time predictions, Design space explorations, Deep learning.
\end{keyword}

\end{frontmatter}


\section{Introduction}

Topology optimization (TO) seeks the optimum material distribution given an objective function, a design domain, and a set of boundary conditions. Each optimization is an iterative process, often requiring hundreds of iterations. From a computational standpoint, running a 3D TO requires the investment of most of the available computational effort to solve the analysis equations for every iteration. The associated computational cost, which is significant for any problem that approaches real conditions, typically depends on the dimensionality and resolution of the domain, on the number of design variables being used, as well as on the numerical solution procedure itself. A good and relatively recent review of the popular numerical solution procedure approaches can be found in \cite{sigmund2013topology} and a detailed discussion of the computational cost involved can be found in \cite{amir2011reducing} and \cite{limkilde2018reducing}. Very recently, various machine learning algorithms have been proposed to tackle the massive computational cost of established gradient-based TO methods, including convolutional neural network (CNN) \cite{banga20183d}, generative adversarial network \cite{li2019non}, and conditional generative adversarial network \cite{yu2019deep}. While these approaches are promising, the training needed by these methods requires large datasets that are computationally expensive to generate, as discussed below. Furthermore, these methods have not been shown to be capable of handling multiple  boundary conditions, domains with different topologies and geometries, or high-resolution 3D domains. 

In this paper we introduce a transfer learning method based on a convolutional neural network that (1) can handle high-resolution 3D design domains of various shapes and topologies; (2) supports real-time design space explorations as the domain and boundary conditions change; (3) requires a much smaller set of high-resolution examples for the improvement of learning in a new task compared to traditional deep learning networks; and (4) is multiple orders of magnitude more efficient than the established gradient-based methods.

\section{Background}

The published literature on topology optimization has exploded over the last two decades to include methods that use shape and topological derivatives or evolutionary algorithms formulated on various geometric representations and parametrizations. Different approaches have been developed to search for the optimal shape, including approaches based on material density functions \cite{bendsoe1989optimal}; level sets \cite{allaire2002level,chen2007shape}; topological derivatives \cite{eschenauer1994bubble,novotny2003topological}; phase fields \cite{bourdin2003design}; and several other variations \cite{sigmund2013topology}. One popular and established approach based on material density, known as  Solid Isotropic Material with Penalization (SIMP) \cite{bendsoe1989optimal}, finds the optimal topology by changing the material density of elements in the unit interval $[0,1]$. The methods based on topological derivatives aim  to  predict  the  sensitivity of the problem to the addition of  an  infinitesimal  hole  at  prescribed locations inside  the  design domain, and this information is used to generate new holes \cite{sokolowski2009topological}. 

One of the key challenges of all these methods is the immense computational cost associated with 3D topology optimization problems, so it is not surprising that there is an extensive body of work dedicated to improving the computational cost of TO. Parallel computing has been employed in \cite{borrvall2001large} for large scale topology optimization by dividing the domain into sub-domains that are independently solved on separate processors. The results presented in \cite{borrvall2001large} present an optimization of a domain with 24x80x128 elements in 234 minutes using 16 processors. The work described in  \cite{wu2015system} relies on a high-performance multigrid GPU solver to find the optimum solution of models with millions of elements. Their method solved on the GPU a 200x100x100 cantilever beam with 50 iterations, and volume fraction of 0.8 in 2.4 minutes. Observe that standard SIMP implementations \cite{andreassen2011efficient,liu2014efficient} require more than 120 iterations for convergence, even for simple cantilever beams. Design space adjustment and refinement is employed in \cite{jang2008design} to speed up the computations for large-scale domains. This work performs TO for an MMB beam with 10800 elements in 5.5 minutes, which is roughly a third of the time required by a SIMP implementation to produce a solution to the same problem. Other papers, such as \cite{kim2012new}, have employed the reduction of the the number of design variables to decrease the computational cost, and showed the capability to produce an optimal solution of a cantilever beam in 10.12 minutes compared to 20.2 minutes required by a SIMP-based solver.

As mentioned above, machine learning methods have been recently used to tackle the computational efficiency of the topology optimization problem. For example, the work discussed in  \cite{lynch2019machine} uses machine learning to tune the numerical parameters that control the convergence of established topology optimization algorithms to avoid the manual tuning, which is computationally costly. 
CNNs have been applied to estimate the optimal topologies for 3D low resolution beams \cite{banga20183d}, or 2D domains \cite{sosnovika2017neural,lin2018investigation,gaymann2019deep,o2019standard}. Moreover, CNNs have been very recently used to predict the optimum 2D structure for simple low resolution 2D domains \cite{zhang2019deep}. In this paper, the authors claim a generalization ability of their network primarily because they explicitly input into the network the initial displacement and strain fields as well as the volume fraction, rather then the explicit boundary conditions. However, they use 80,000 training samples and 10,000 test samples, which is simply prohibitive for any problem that is of reasonable complexity. At the same time, different versions of GANs have been used to optimize the topology of 3D cantilever beams \cite{rawat2019application}, 2D low resolution cantilever beams \cite{rawat2019novel,shen2019new}, and 2D high resolution beams \cite{li2019non,yu2019deep}. Variational auto-encoders (VAE) and supported vector regression (SVR) have been applied to 2D low resolution domains with different boundary conditions \cite{guo2018indirect}, and cantilever beam \cite{lei2019machine}, respectively. All these methods can reduce the TO computational time for given domain resolution, set of boundary conditions, and initial domain as long as large amounts of data are available. However, applying these methods to tasks that the algorithms have not been trained on requires new large sets of training data. This is not only impractical for any topology optimization problem of reasonable complexity, but also makes these methods unsuitable for design space explorations. 

A key assumption of most machine learning algorithms is that the training data and task data are in the same feature space and have the same distribution. This is a reasonable assumption as long as the training data is relatively painless to generate and abundant. However, this is definitely not the case in many engineering application, including topology optimization. Transfer learning has emerged as one promising learning algorithm that has the potential to greatly improve the learning performance by limiting the amount of training that needs to be performed to adapt the algorithms to new scenarios.  It aims to imitate one of the distinctive features of human intelligence, that is, to effectively transfer previously learned knowledge to new domains \cite{weiss2016survey, torrey2010transfer}. Transfer learning has been successfully used in medical image processing \cite{khatami2018sequential}, and brain-computer interface calibration \cite{hossain2018multiclass}. 

\begin{figure}[t]
	\centering
	\includegraphics[width=0.8\Columnwidth]{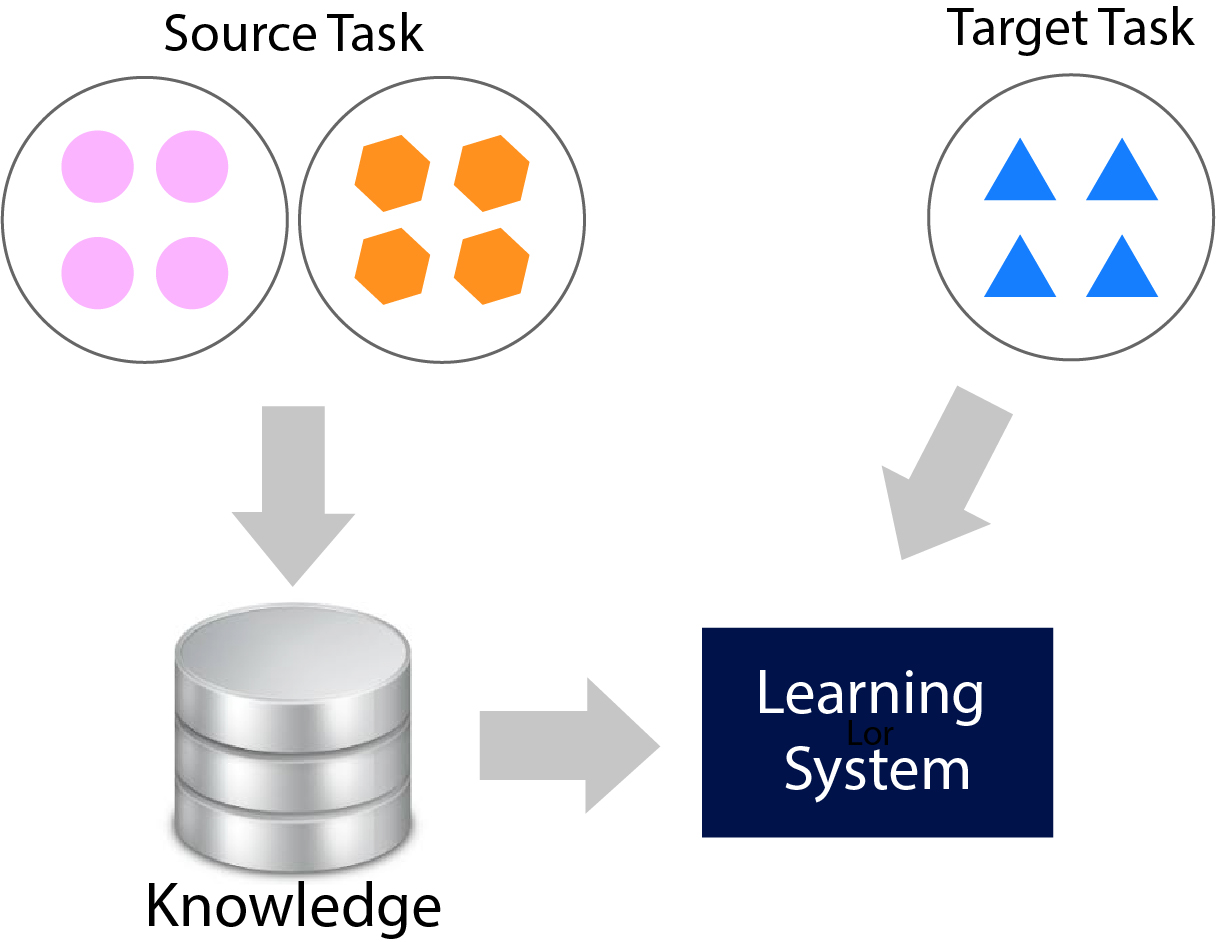}
	\caption{A schematic diagram of transfer learning after \cite{Pan2010survey}}
\label{fig1}
\end{figure}

\subsection{Contributions and Outline}

In this paper we propose a transfer learning approach that we developed specifically for topology optimization. We show that the method produces highly accurate predictions of the optimal 3D topologies at real-time rates for non-trivial 2D and 3D high resolution TO problems. Furthermore, we show that the proposed method serves as the first practical underlying framework for real-time 3D design explorations based on topology optimization, and that fine tuning/retraining the proposed learning algorithm for new tasks can be done with a much smaller high-resolution dataset than the traditional deep learning networks. To the best of our knowledge, this paper documents the first promising attempt to use transfer learning for topology optimization. 

The rest of the paper is organized as follows. Section \ref{formulation:sec} presents the formulation of the proposed method, as well as a detailed description of the implementation and data generation. This section also includes a discussion of the source and target networks as well as of the two metrics that we used to evaluate our network. Section \ref{results} illustrates the generality and flexibility of our approach by providing a variety of 2D and 3D examples with different resolutions, boundary conditions and design spaces, including design domains that are \textit{unseeen} to the source network. These examples show that the proposed method supports real-time design space explorations as the domain and boundary conditions change and is multiple orders of magnitude more efficient than the established methods for both 2D and 3D design scenarios.  Finally, Section \ref{conclusions:sec} summarizes the key advantages and limitations of the proposed method and of its potential applications.

\begin{figure*}[t]
	\centering
     \includegraphics[height=4cm]{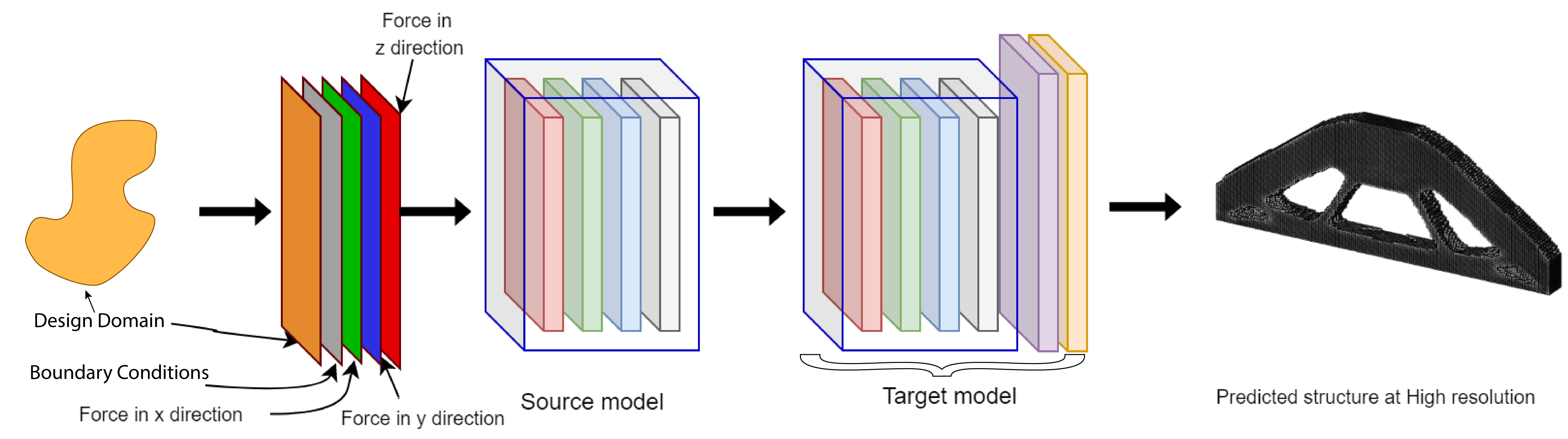}
     \caption{A diagram of the proposed method for 3D topology optimization.}
     \label{fig:2}
\end{figure*}

\section{Problem Formulation}\label{formulation:sec}
We describe the proposed method in the context of the well known minimum compliance topology optimization problem, which is considered to be a ``global response'' of a structure. However, our method can be equally well applied to TO problems that consider more local responses in their objective functions, such as stress-based TO \cite{le2010stress}. 

The general topology optimization method aims to find the spatial distribution of material $\rho(\mathbf{x})$ that minimizes an objective function $f(\Omega, \rho)$, subject to various constraints $g_i \leq 0$, where the state field $\mathbf{u}$ satisfies given state equation. It is common to assume that the objective function is expressed as an integral over a local function such as the strain energy density \cite{sigmund2013topology}, and to solve the problem using the finite element method. In the SIMP method, each element is associated with a density variable $\rho_{e} \in [0,1]$, where $0$ corresponds to an empty element, and $1$ to an element completely filled with material. This optimization problem can be formulated as 

\begin{eqnarray}\label{eq.1}
   min &:& f(\Omega, \mathbf{\rho}) = \mathbf{U^{t}KU} = \sum E_{e}(\rho_{e})\mathbf{u}_{e}^{t}\mathbf{k}^0_{e}\mathbf{u}_{e} \\
    s.t  &:& \mathbf{KU} = \mathbf{F} \\
    		 & & \sum \rho_{e}v_{e} \leq V_{max} \\  
         & &  0\leq \rho \leq 1
\end{eqnarray}
where $\Omega$ is the domain; $\mathbf{\rho}$ is the design variable vector of densities; $\mathbf{U}$ and $\mathbf{F}$ are the global displacement and force vectors, respectively; $\mathbf{K}$ is the global stiffness matrix; $\mathbf{u}_{e}$ is the element displacement vector; $\mathbf{k}^0_{e}$ is the element stiffness matrix for an element with unit Young's modulus; $v_{e}$ and $\rho_{e}$ are the element volume and density of element $e$, respectively; and $V_{max}$ is the volume upper bound. $E_{e}(\rho_{e})$ is the element's Young's modulus determined by the element density $\rho_{e}$:
\begin{equation}
    E_{e}(\rho_{e}) = E_{min} + \rho_{e}^{p}(E_{0}-E_{min}), \hspace{1cm}  \rho_{e} \in[0,1] \label{eq.2}
\end{equation}
In equation (\ref{eq.2}), $E_{0}$ represents the stiffness of the material, $E_{min}$ is a very small number, and $p$ is a penalization factor \cite{sigmund2007morphology,sigmund2013topology}.

\subsection{Topology Optimization with Transfer Learning}
One of the distinctive features of human intelligence is the ability to effectively transfer previously learned knowledge to new application domains. We all use this capability every single day even without realizing it \cite{Steiner2001}. In contrast, most machine learning algorithms are trained on and function only on well defined tasks. Transfer learning aims to improve this limitation by transferring the knowledge from a source task to improve the performance in target task that is different from but related to the source task. Every transfer learning algorithm uses specific learning algorithms to learn the tasks at hand, which is why transfer learning is often described as extensions of those learning algorithms.

In a broad sense, transfer learning deals with 3 different questions:  (1) \textit{what} information should be transferred; (2) \textit{how} to transfer the information;  and (3) \textit{when} to transfer it. As an example, for the topology optimization task discussed in this paper, we transfer the weights and biases of all layers of the source network except for the last layer. To address the second question, we implemented a mechanism to transfer this knowledge from the source network to the target network with minimal retraining. The third question deals with establishing the cases when the knowledge transfer should be performed, i.e., the validity of the transfer learning method \cite{Pan2010survey}, and this is often addressed by measuring the performance to new tasks. This is also the approach that we take in this paper.

Because optimizing a high resolution domain using  state of the art gradient based TO methods, including the popular SIMP method, is always computationally very expensive,  generating sufficient training data for practical design scenarios becomes a crucial bottleneck. This is why we developed a deep transfer learning method that uses a fully convolutional neural network, which allows us to transfer the knowledge obtained from training the algorithm on low resolution models to be usable on high resolution cases with different design domains and boundary conditions that the source network has not been trained on.

Figure \ref{fig1} illustrates how transfer learning works. First, a source model is built and trained with a large amount of low resolution data,  which is relatively inexpensive from a computational point of view. This is followed by transferring what the source model learned to a target model operating on different but related tasks. This process allows the use of a much smaller amount of data to retrain/fine tune the target model to improve learning in the target task. In our implementation, the source and target models are CNN based decoder-encoders with the source model trained on large amount of low resolution (and thus relatively inexpensive) data, and the target model retrained for the new task with relatively small high-resolution datasets.
 
The overall architecture is illustrated in Fig.\ref{fig:2}. The input is the design space and the boundary conditions. For the examples presented in this paper we only considered externally applied forces, but adding externally applied torques is straightforward. During the final step, we build the target network described in section \ref{networks} by augmenting the source network with additional layers, and train it with the high resolution data.

\subsection{Network Architecture and Network Training}\label{networks}

Convolutional Neural Networks (CNNs) have been successfully used in object classification and segmentation tasks. More recently, a number of papers have shown CNNs to be trainable and effective on large datasets for solving inverse problems in imaging, such as denoising, deconvolution, super-resolution, and medical image reconstruction. These inverse problems practically involve the determination of an image from noisy measurements. Since the datasets output by SIMP can loosely be considered as a special case of these inverse problems,  the architectures of our source and target networks are inspired by some of the work described in reviewed in \cite{mccann2017convolutional} and elsewhere.

\begin{figure*}[h!]
	\centering
	\includegraphics[height=8cm]{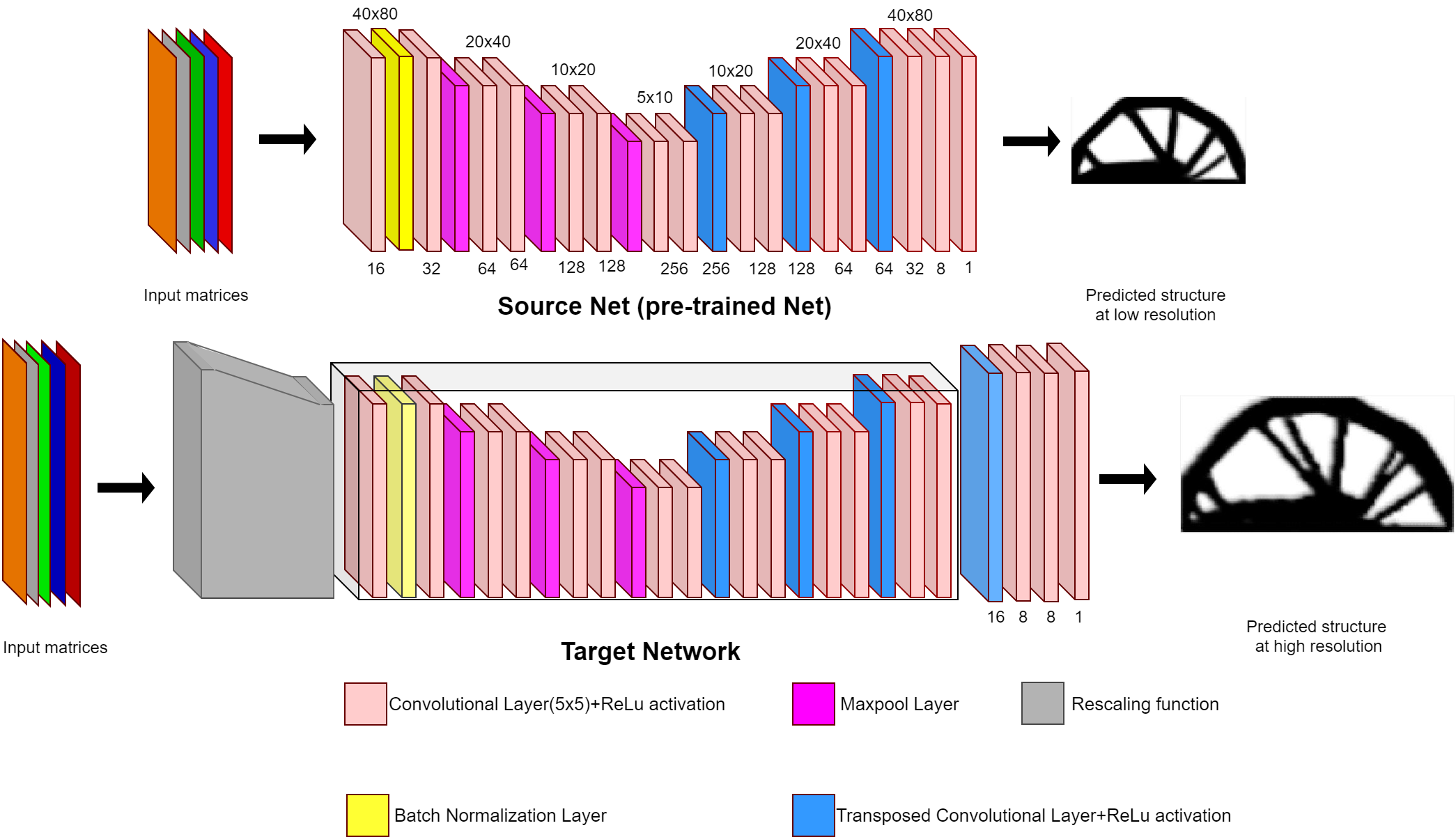}
    \caption{The architecture of the source and target networks. Numbers below the boxes denote the number of filters used.}
    \label{fig:4}
\end{figure*}

Our source network is a two-dimensional encoder-decoder based convolutional neural network (CNN) illustrated in Fig.\ref{fig:4}. The encoder part includes eight convolutional layers with the rectified linear unit (ReLU) activation function, and three max-pooling layers. The decoder part of our source network includes three transposed convolutional layers and seven convolutional layers using the ReLU activation function. We trained the  network using the  ADAM optimizer \cite{lehman2010revising}, which is a standard gradient-based optimization algorithm. The ADAM optimizer finds the optimal weights and biases of the network that minimize the loss between the predicted structures and the simulated structures according to the mean squared error (MSE).


We built our target network on top of the source network as illustrated in Figure \ref{fig:4}. To the end of the source network we added one transposed convolutional layer to increase the output dimension of the pre-trained network to the higher resolution, as well as three trainable convolutional layers. To the front of the network, we added a rescaling function to downsample the higher resolution input to the target network to the lower resolution required by the source network.

Prior to training the target network, we remove the last layer of the pre-trained network. This modification is based on measuring the accuracy of the predictions as described in the next section. This modified target network is trained as described in section \ref{results} with ADAM as the optimizer and MSE as the loss function. 

Importantly, we trained the source network only once for a given dimension of the space (2D or 3D). This trained source network is integrated within the target network, which is retrained/fine tuned with smaller datasets for new design domains and/or boundary conditions. 

\subsection{Data Generation}

We used freely available SIMP-based topology optimization finite element codes \cite{andreassen2011efficient,liu2014efficient} to generate our training and test cases, and we modified the codes to automate the training/test case generation for different domains and boundary conditions. The input to the codes are the following design variables and physical quantities: voxelized domain geometry,  volume fraction, filter radius (see  \cite{sigmund2007morphology,bourdin2001filters} for details), loading boundary conditions (number, magnitudes, directions, spatial locations), and displacement boundary conditions. We prescribed to the SIMP codes a volume fraction equal to 0.5 and a 1.5 filter radius.

\begin{figure*}[t]
	\centering
	\includegraphics[width=2\Columnwidth]{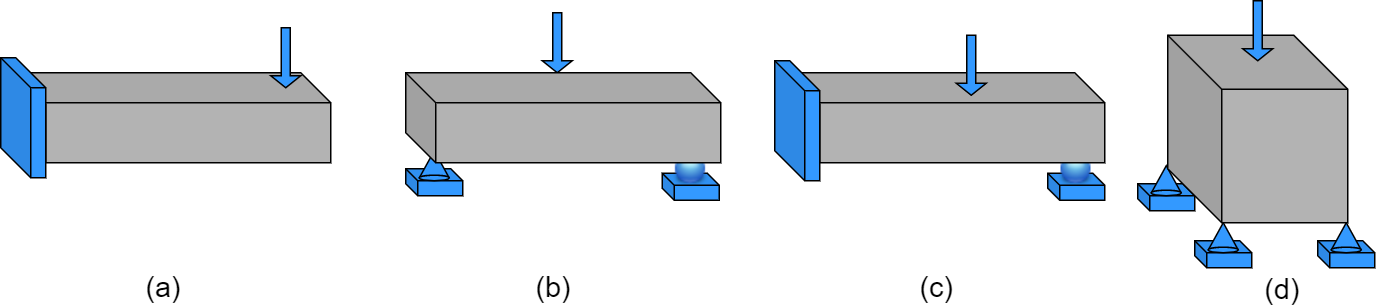}
     \caption{The randomly selected displacement boundary constraint cases: (a) Cantilever Beam, (b) Simply Supported Beam, (c) Constrained Cantilever, and (d) The boundary conditions for the cases shown in Figures \ref{fig:cube} and \ref{fig:cubesph}.}
     \label{fig:3}
\end{figure*}

We sampled the magnitudes of the force components using uniform random sampling within the range $[-100,100]$N. The spatial location on the domain boundary where the external load was applied was chosen based on a uniform random sampling within prescribed ranges along the coordinate axes. For example, the force components $P_x$ and $P_y$ for a 2D beam domain are applied within the range $[\frac{b_x}{2},b_x]$ and $[0, b_y]$, respectively, where $b_x$ and $b_y$ are the beam dimensions in the $x$ and $y$ directions. We also used a discrete random sampling to select one of the defined displacement boundary constraints illustrated in Fig.\ref{fig:3}.

With this random sampling of the boundary conditions we generated two datasets: a low-resolution dataset for our source network, and a smaller, high resolution dataset  for the target model. Those two datasets with distinct elements, were then split into training and testing datasets, where the latter was at least 20\% of the former. Details about the sizes of these datasets are provided in Sections \ref{2DStructures} and \ref{3DStructures}. The individual sizes of the testing datasets for our examples are shown in Tables \ref{tab:2Daccuracies} and \ref{tab:3Daccuracies}. 

We use matrices to store the data fed into the five channels to our network architecture. For example, we used five channels for the 2D cases, as follows:
\begin{enumerate}
    \item First channel: Initial density value for each voxel.
    \item Second channel: Constraints in the $x$ direction initialized to zero, then the elements corresponding to the constrained elements are set to 1.
    \item Third channel: Constraints in the $y$ direction initialized to zero, then the elements corresponding to the constrained elements are set to 1.
    \item Fourth channel: Force value in $x$ direction at each voxel,
    \item Fifth channel: Force value in $y$ direction at each voxel.
\end{enumerate}

\subsection{Evaluating the Network}\label{net:eval:sec}
We employed the following two criteria for the evaluation of our method against state of the art methods:
\begin{enumerate}
    \item \textit{Mean Squared Error} (MSE), which is defined as: 
    \[ MSE = \frac{\sum_{j=1}^{M}\sum_{i=1}^{N} (y_{pred}^{ij} - y_{true}^{ij})^2}{N \cdot M} \]    
    where $N$ and $M$ are the number of rows and columns, $y_{pred}^{ij}$ and $y_{true}^{ij}$ are the predicted and reference values of element located in the $i^{th}$ row and $j^{th}$ column, respectively. 
    \item \textit{Binary Accuracy} (BA), a widely used measure that compares the binarized value of predicted elements with the actual value \cite{christen2007quality}:
    \[BA = \frac{TP + TN}{N}\]
    where $TP$ (True Positive) is the number of elements correctly predicted as 1, $TN$ (True Negative) is the number of elements  correctly predicted as 0, and $N$ is again the total number of elements. Prior to calculating the binary accuracy, we rounded the element values to the nearest integer, that is, either to 0 or 1. 
    
\end{enumerate}

As the average of the squared deviations, the Mean Squared Error is a second sample moment about the origin of the error, and is the minimum variance unbiased estimator \cite{birch1983classroom}. For a given predictor, the closer MSE is to zero, the better the predictor performance. Furthermore,  accuracy reflects the overall ratio of correct predictions. A common version of this measure, namely the binary accuracy, uses binarized density values in the ground-truth and predicted domains, and shows how accurately the network predicts the existence of material in each voxel. The values of the binary accuracy lie within the unit interval, and the closer the binary accuracy is to 1, the better the prediction. In this work, we use the binary accuracy to measure how accurate our network is in predicting the existence of material, and we use MSE to estimate\footnote{These two evaluation metrics do not always agree. For example, assume that the predicted and actual values are 0.49 and 0.51 respectively. MSE tells us that the error is 0.04\% which means the prediction is very accurate, But binary accuracy treats the predicted value as 0 and the actual value as 1, and implies a 0 binary accuracy. A more detailed discussion can be found in standard texts on probability.}  how close are the density predictions to the ground-truth values. Our experiments achieved an average binary accuracy and MSE around 95\% and 3\%, respectively. All predictions were performed on a Dell Intel Xeon Processor E5-2650 v3 CPU with 64 GB RAM and Nvidia Quadro K2200 4GB GPU. All training and test cases were generated on the UConn HPC facility running Red Hat RHEL7 operating system.

For the optimal topologies predicted by our algorithms, we are also providing the corresponding compliance error relative to the compliance of the ground truth optimal structures. Since our predicted structures are directly output by our algorithms without any post-processing/beautification steps, the resulting compliance errors shown for the examples are particularly promising.

\section{Results}\label{results}

\subsection{2D Structures}\label{2DStructures}

\subsubsection*{Comparison with Ground Truth (SIMP)} 

We used a freely available Matlab Code \cite{andreassen2011efficient} to generate the ground truth results. For the 2D examples, we trained the source network with 8,000 low resolution cases (40 x 80), and fine-tuned the target model (various resolutions, as shown in Table \ref{tab:2Daccuracies}) with 1500 high resolution cases. 

Importantly, the source network is trained only once and the learned information is being reused. The high resolution cases used for fine tuning the target network have the same resolution as the test cases used in our examples and shown in Table \ref{tab:2Daccuracies}. To generate the training data for all 2D examples,  boundary conditions have been randomly selected from one of the 3 cases shown in Figure \ref{fig:3}(a-c), and the location, orientation and magnitude of the external force have also been randomized. 

We first compared our approach with the performance of the SIMP solver \cite{andreassen2011efficient}  for 2D structures in terms of the output accuracy as described in section \ref{net:eval:sec}, and the average time required to obtain the optimal topology. For this evaluation, we used 2,000 low resolution test cases for the source network and a smaller set of high resolution cases with randomly generated boundary conditions, as summarized in Table \ref{tab:2Daccuracies}. 

\begin{figure*}[p!]
    \begin{subfigure}[b]{.48\textwidth}
    \centering
      \includegraphics[width=.5\linewidth]{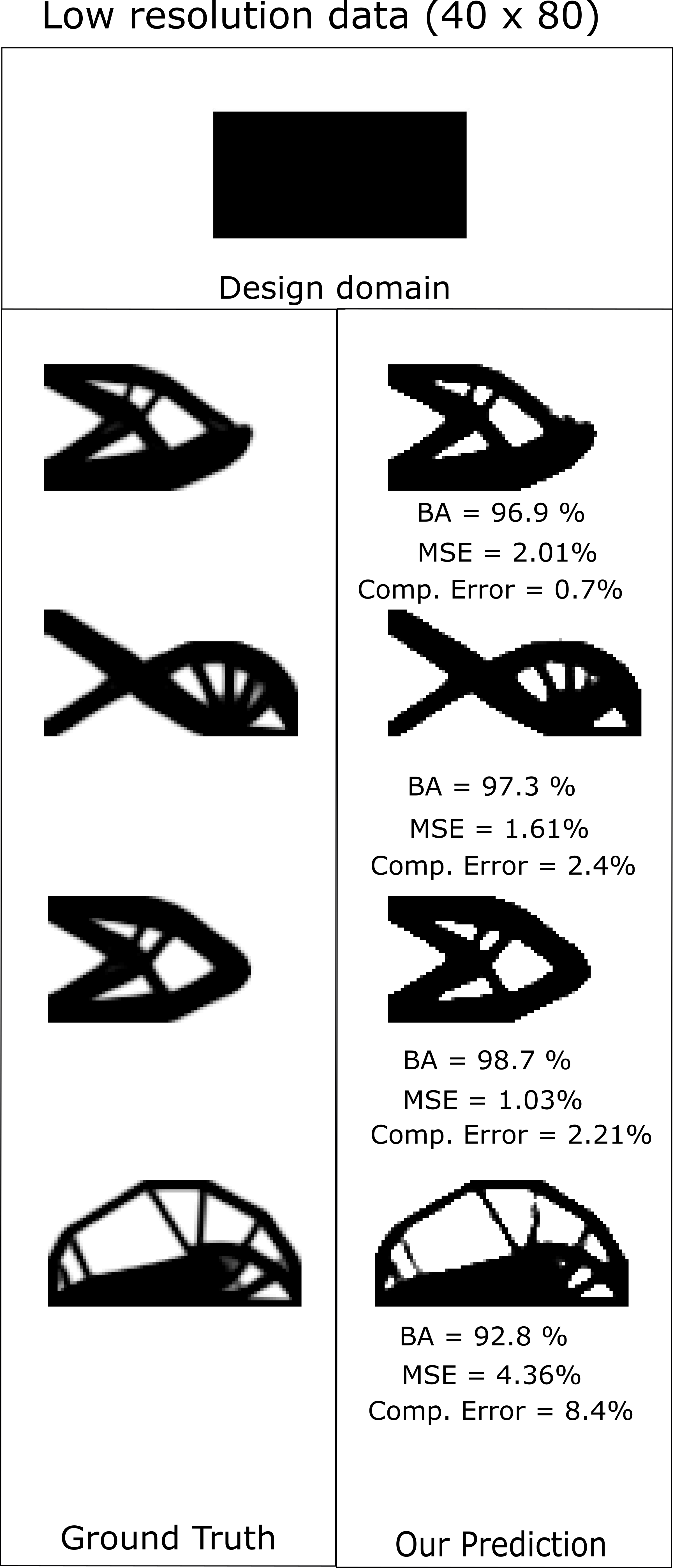}
      \caption{}
      \label{fig:2Dbeam-recta}
    \end{subfigure}
\hspace{-4.3cm}
    \begin{subfigure}[b]{.48\textwidth}
    \centering
      \includegraphics[width=.5\linewidth]{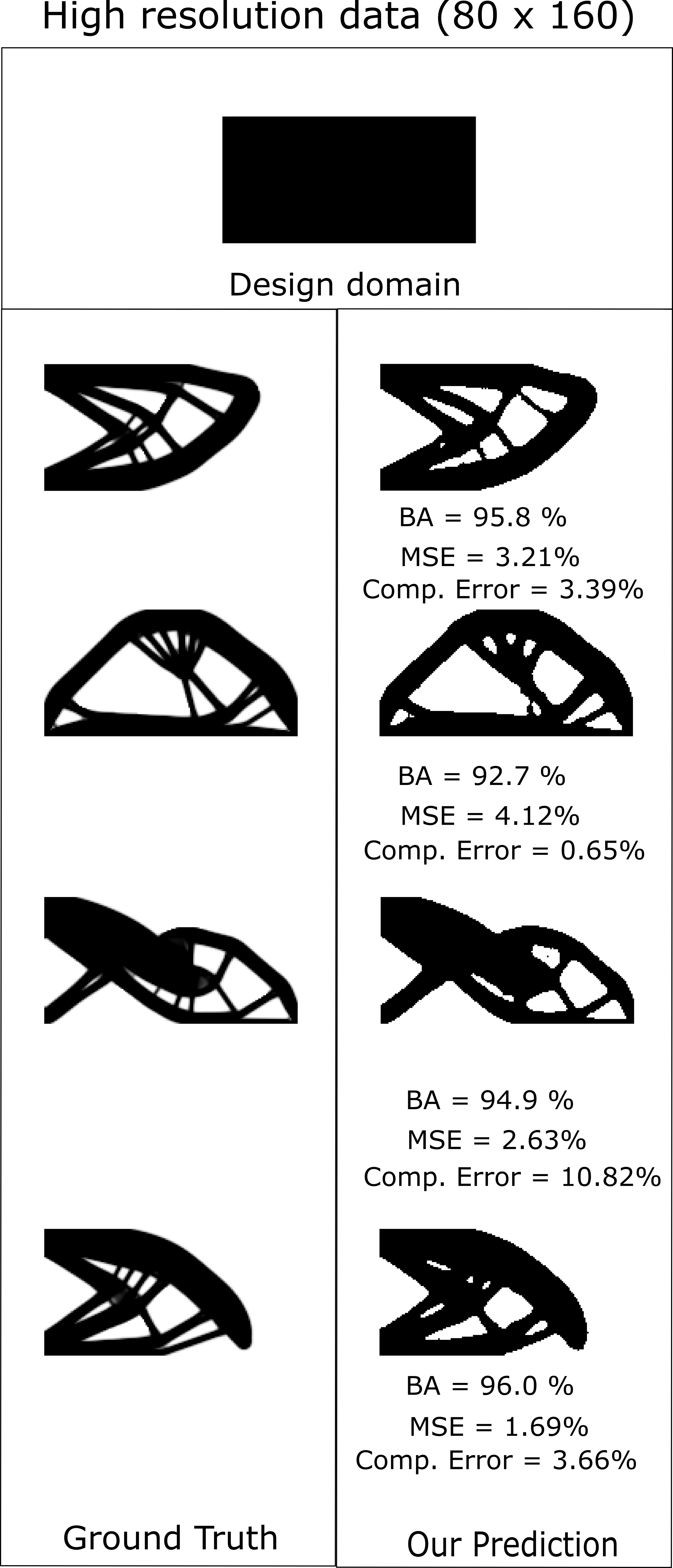}
      \caption{}
      \label{fig:2Dbeam-rectb}
    \end{subfigure}
\hspace{-4.3cm}
    \begin{subfigure}[b]{.48\textwidth}
    \centering
      \includegraphics[width=.5\linewidth]{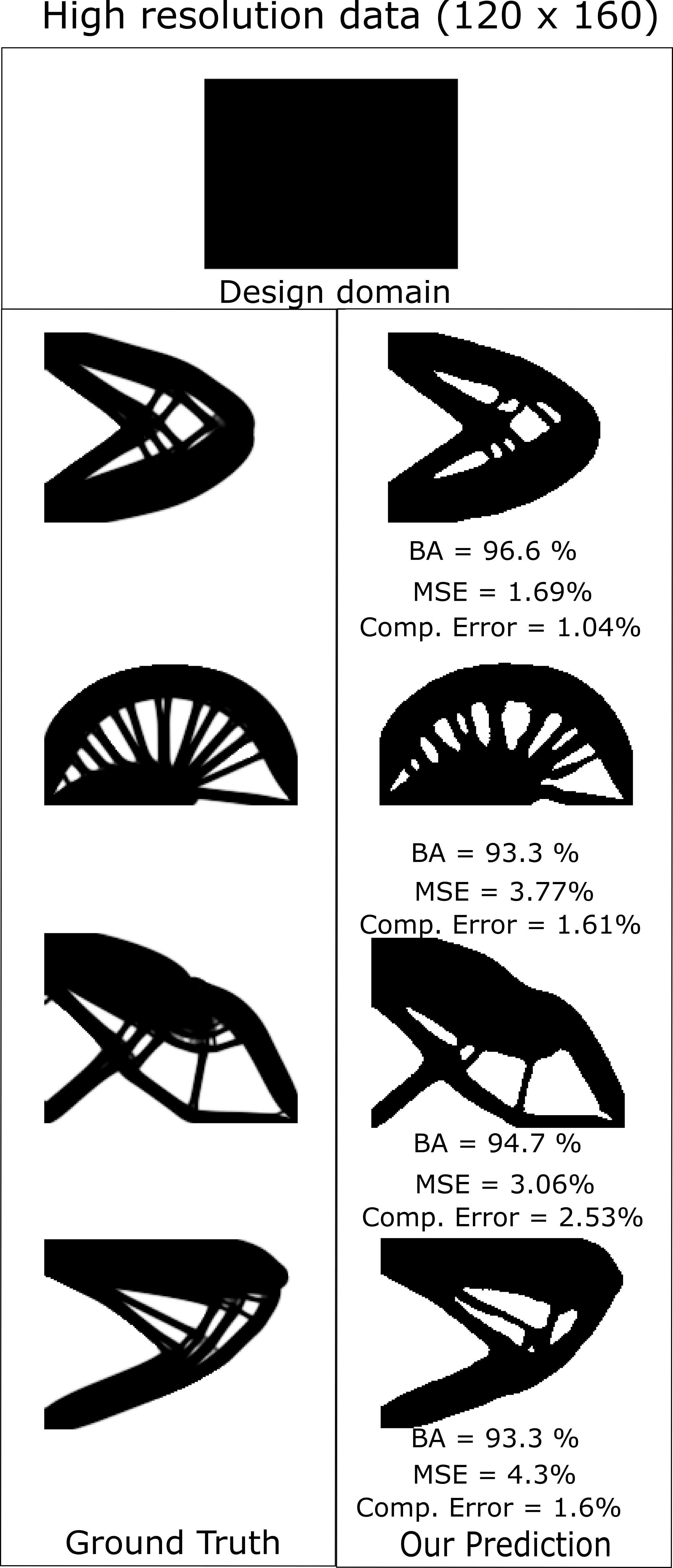}
      \caption{}
      \label{fig:2Dbeam-rectc}
    \end{subfigure}
    \begin{subfigure}[b]{.48\textwidth}
    \centering
      \includegraphics[width=.5\linewidth]{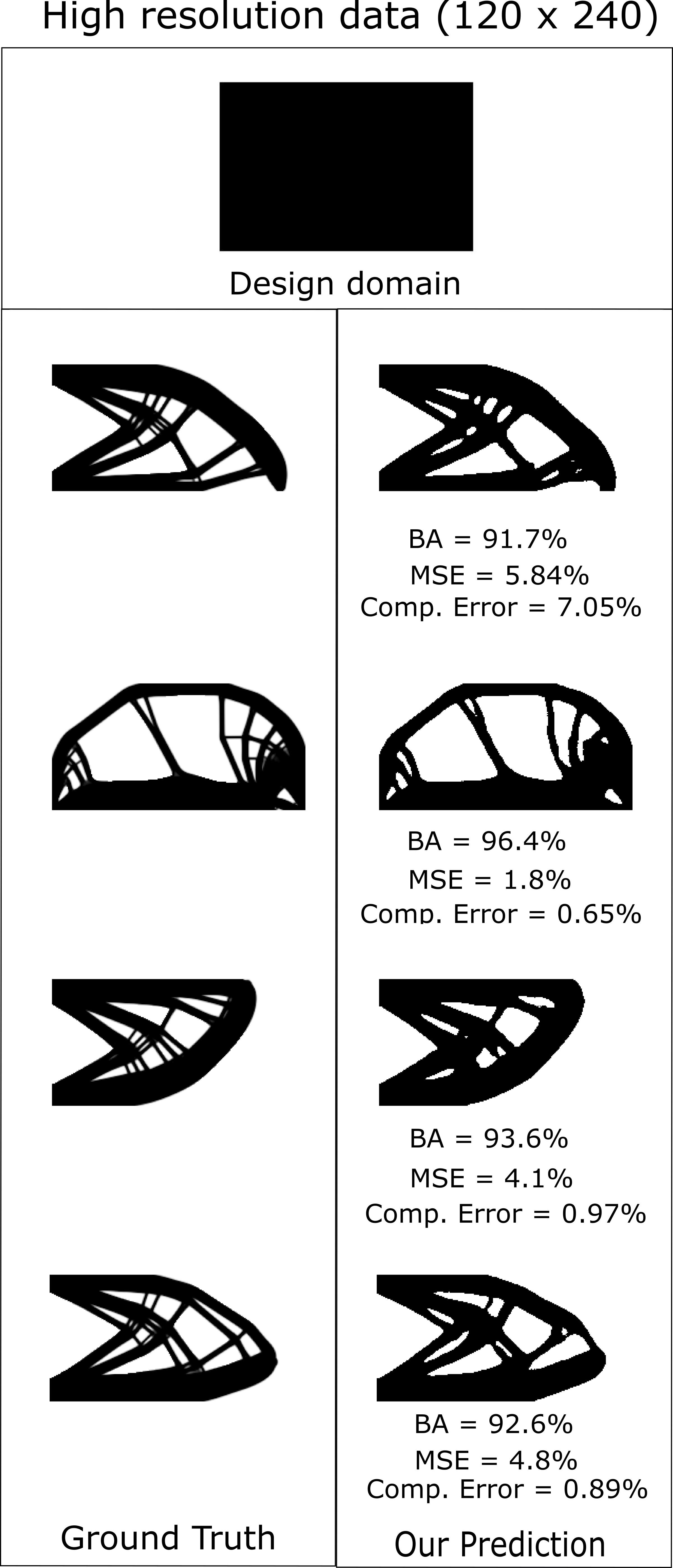}
      \caption{}
      \label{fig:2Dbeam-rectd}
    \end{subfigure}
\hspace{-4.3cm}
    \begin{subfigure}[b]{.48\textwidth}
    \centering
      \includegraphics[width=.5\linewidth]{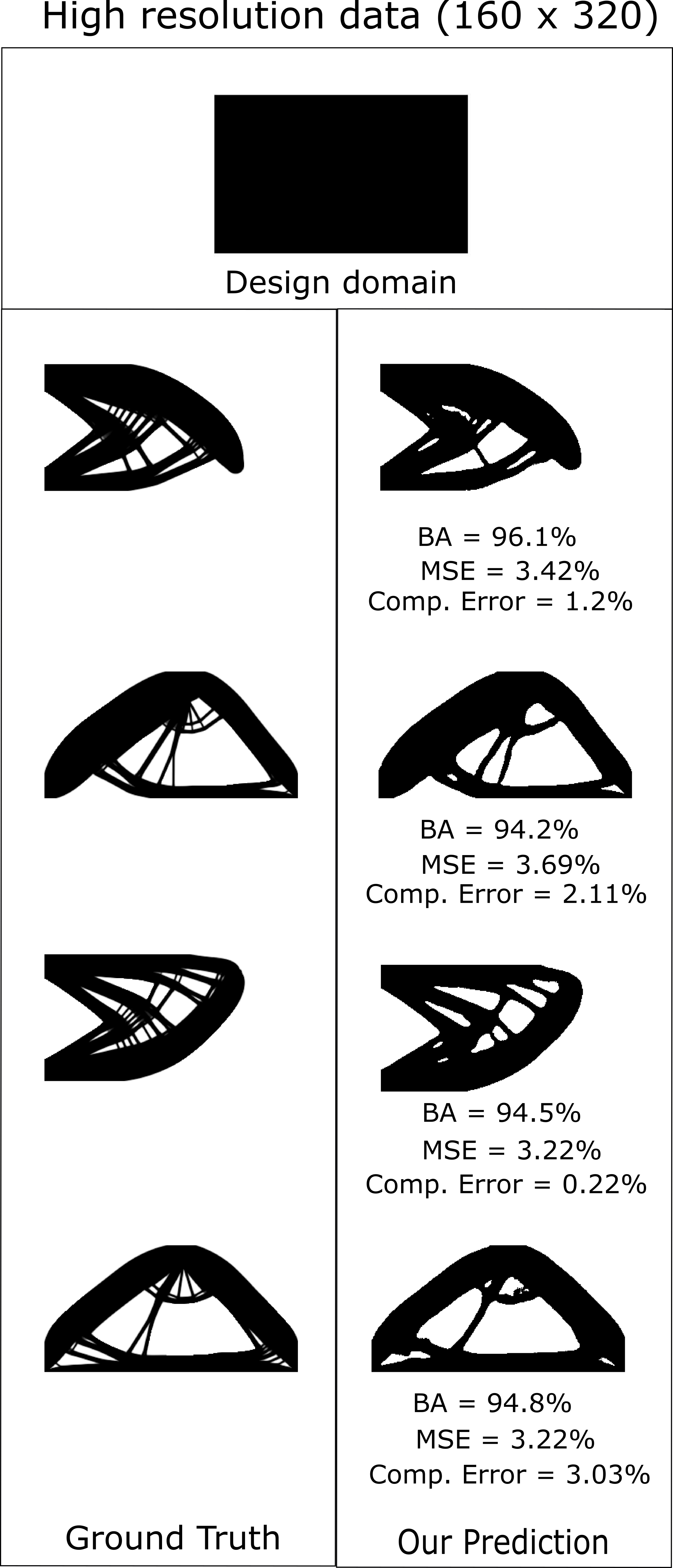}
      \caption{}
      \label{fig:2Dbeam-recte}
    \end{subfigure}
\hspace{-4.3cm}
    \begin{subfigure}[b]{.48\textwidth}
    \centering
      \includegraphics[width=.5\linewidth]{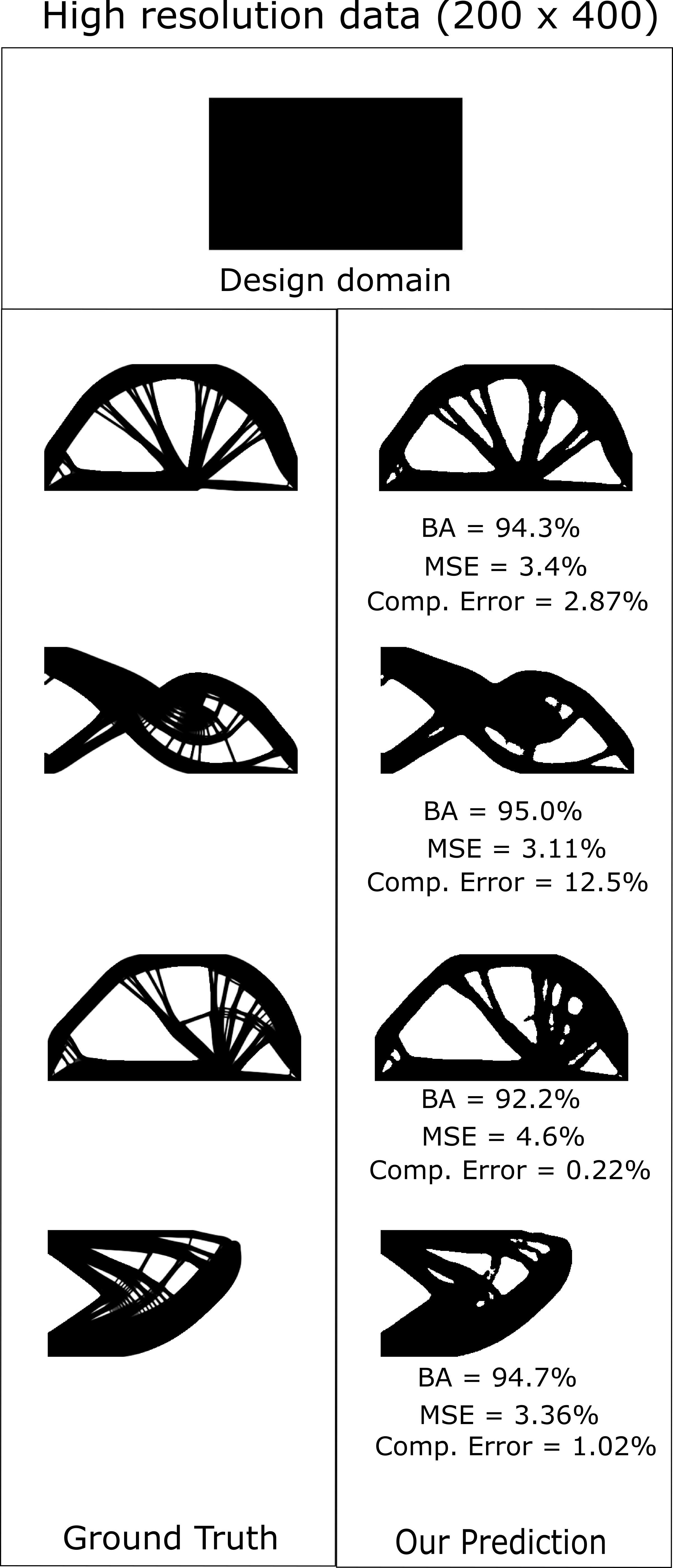}
      \caption{}
      \label{fig:2Dbeam-rectf}
    \end{subfigure}
\caption{Comparison between the ground truth (SIMP optimized) 2D structures and our predictions of the optimal structures. Fig. (a) shows the prediction of our source network alone. Figs (b-f) show the prediction of the optimal structures output by the fine tuned target model. The individual quality metrics for our predictions are presented in Table \ref{tab:2Daccuracies}, and the prediction time is shown in Table \ref{tab:predictiontime}.}
\label{fig:2Dbeam-rect}

\end{figure*}

Fig.\ref{fig:2Dbeam-rect} shows a side-by-side comparison between the optimal topology output by our method relative to the corresponding ground-truth result (i.e., SIMP-based optimal structures) for different domains and boundary conditions. Specifically, Figure \ref{fig:2Dbeam-recta} shows the optimal 2D  structure output by our source network alone on a low resolution design domain as well as the corresponding MSE, binary accuracy and compliance error. Moreover, Figures \ref{fig:2Dbeam-rect}(b-f) show the optimal 2D structures output by our target network for different domains and boundary conditions, and for design domains with higher resolutions. The predicted and ground truth solutions are not only visually similar, but the corresponding average MSE and average binary accuracy are around 3.6\% and 95\%, respectively, and the resulting compliance error is around 8.7\%. 

\begin{figure*}[ht!]
\centering
    \begin{subfigure}[t]{.5\textwidth}
    \centering
      \includegraphics[width=\Columnwidth]{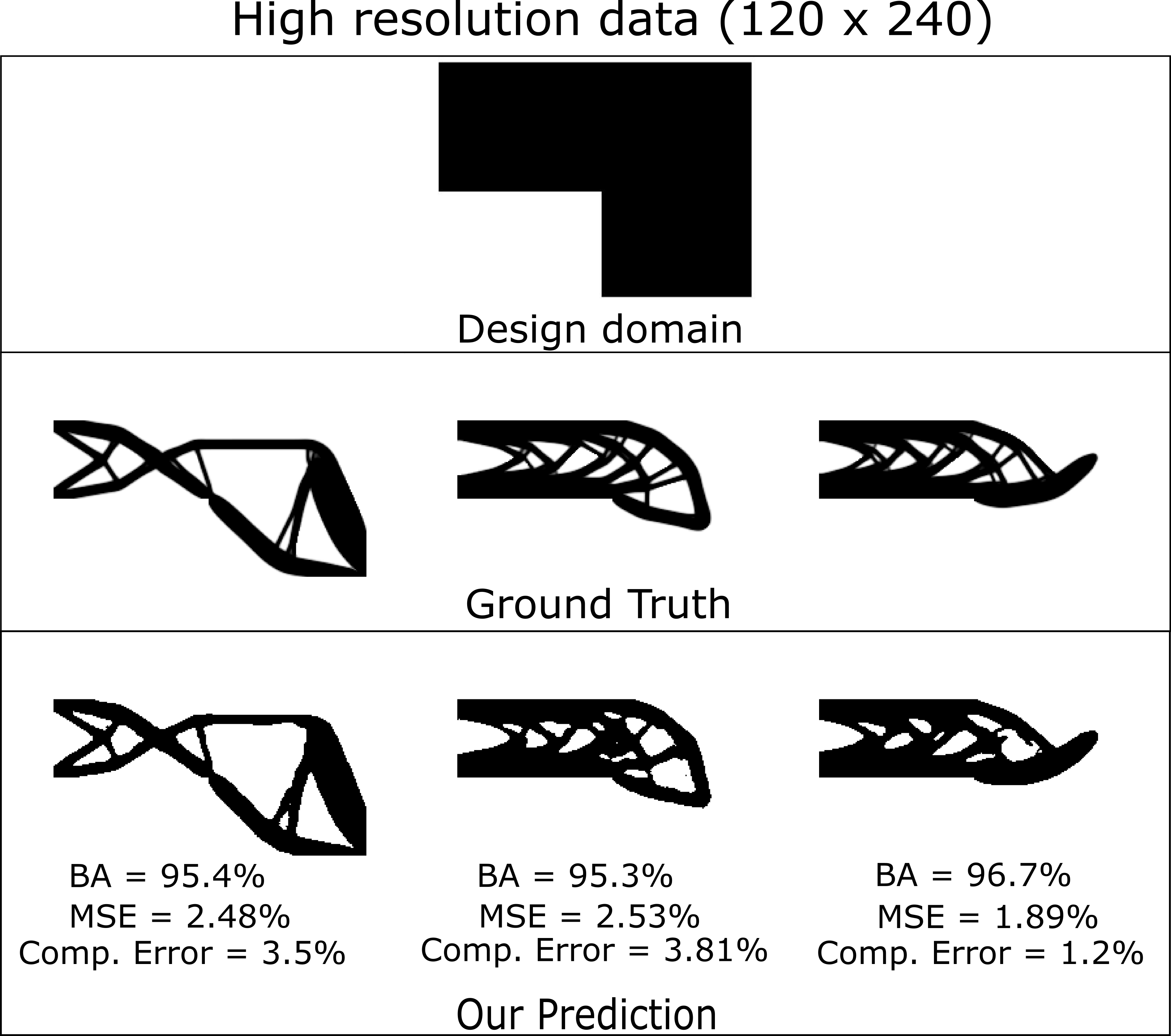}
      \caption{}
      \label{fig:7a}
    \end{subfigure}
    \hspace{-1cm}
    \begin{subfigure}[t]{.5\textwidth}
    \centering
      \includegraphics[width=\Columnwidth]{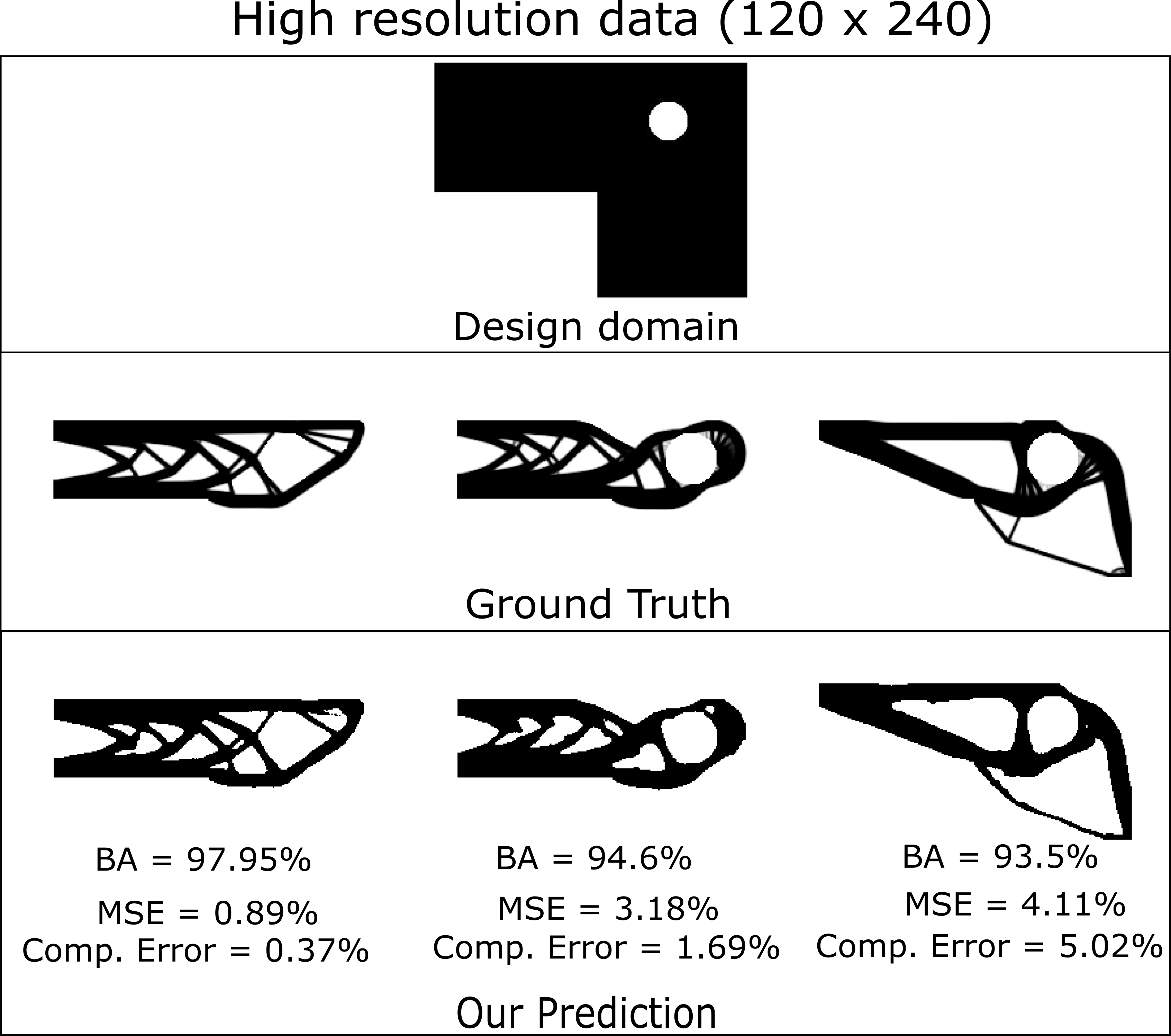}
      \caption{}
      \label{fig:7b}
    \end{subfigure} 
 \caption{Predicted optimal structures versus ground truth (SIMP optimized) for high resolution domains: (a) an L-shaped domain of genus 0, and (b)  L-shaped domain with a hole (genus 1).  The individual quality metrics for our predictions are presented in Table \ref{tab:2Daccuracies}, and the prediction time is shown in Table \ref{tab:predictiontime}. Observe that the source network for the 2D examples has only been trained on the domains shown in Figure \ref{fig:2Dbeam-rect}, so the domains shown in this Figure are \textit{unseen} to the source network. }
\label{fig:7}
\end{figure*}

\begin{figure*}[ht!]
\centering
    \begin{subfigure}[t]{.5\textwidth}
    \centering
      \includegraphics[width=\Columnwidth]{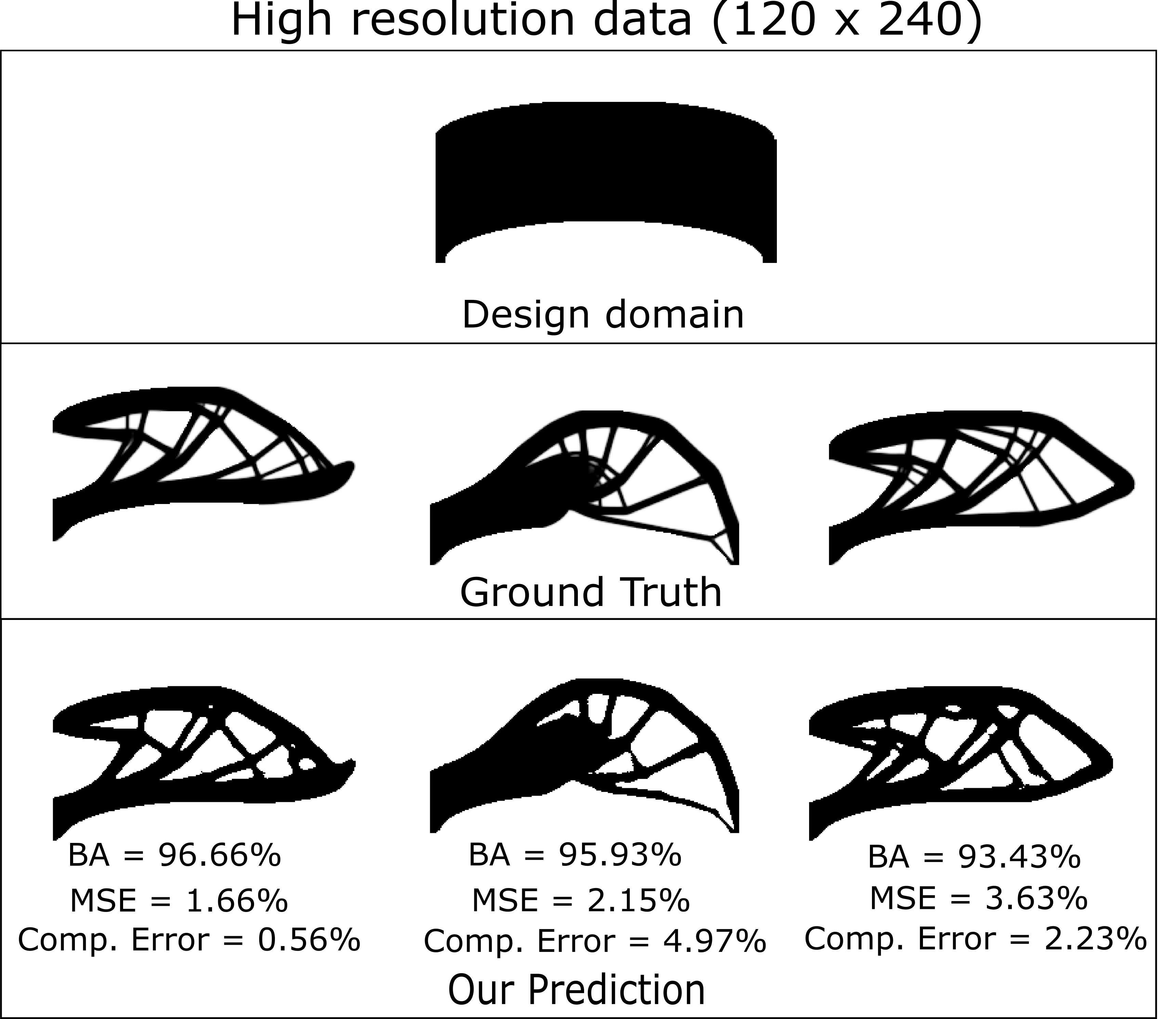}
      \caption{}
      \label{fig:9a}
    \end{subfigure}
    \hspace{-0.3cm}
    \begin{subfigure}[t]{.5\textwidth}
    \centering
      \includegraphics[width=\Columnwidth]{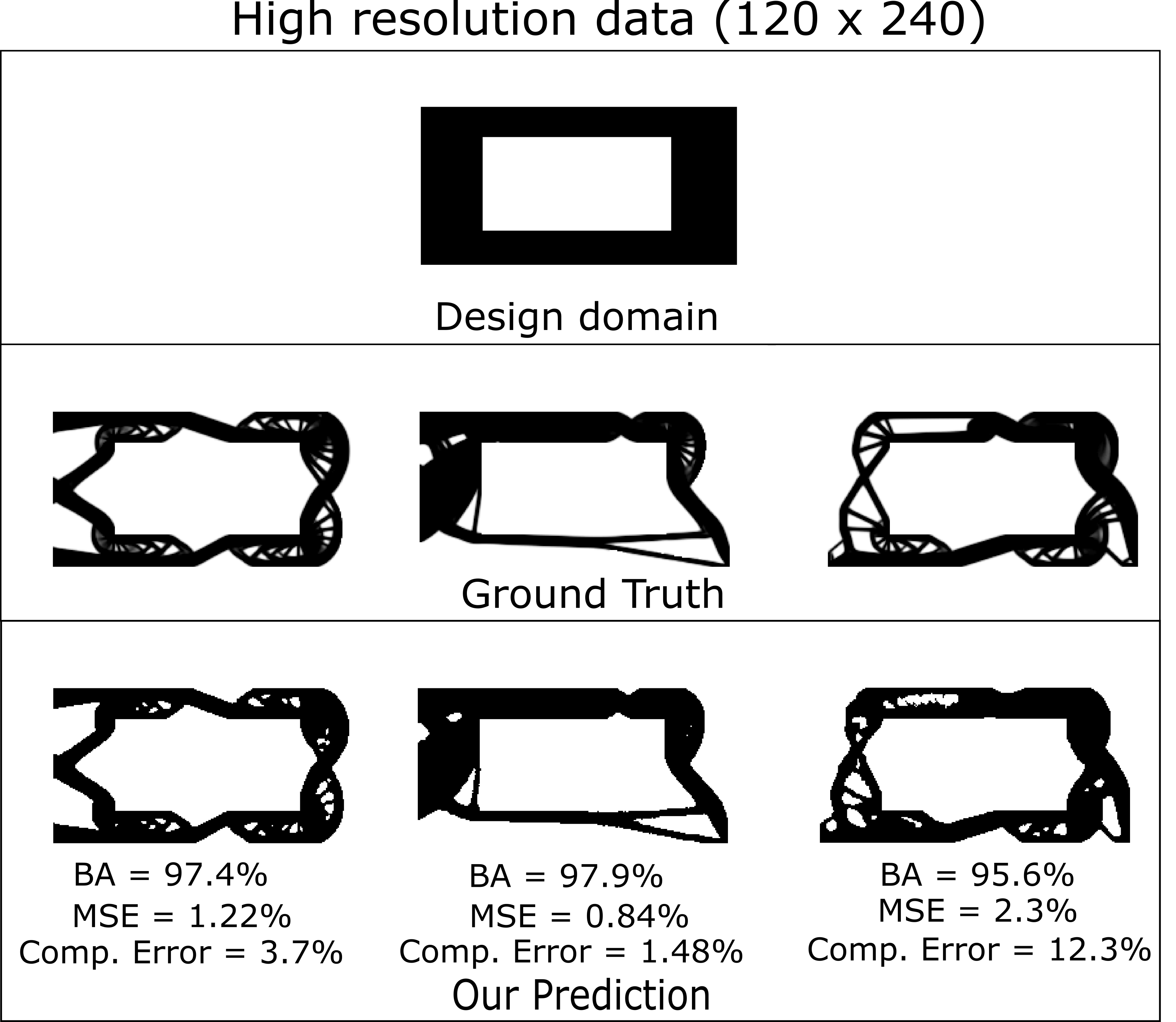}
      \caption{}
      \label{fig:9b}
    \end{subfigure} 
 \caption{Predicted optimal structures versus ground truth (SIMP optimized) for high resolution domains that have different geometry and topology.  The individual quality metrics for our predictions are presented in Table \ref{tab:2Daccuracies}, and the prediction time is shown in Table \ref{tab:predictiontime}. Observe that the source network for the 2D examples has only been trained on the domains shown in Figure \ref{fig:2Dbeam-rect}, so the domains shown in this Figure are \textit{unseen} to the source network. }
\label{fig:9}
\end{figure*}

Figures \ref{fig:7} and \ref{fig:9} show a similar comparison for domains that have different geometry, topology and boundary conditions that the source network has \textit{not} been trained on. The examples shown in Figures \ref{fig:7} and \ref{fig:9} also use different volume fractions (i.e.,0.3 and 0.4, respectively). Importantly, these Figures illustrate not only the performance of our predictions, but also the fact that the same source network can be used to build different target models for substantially different design spaces and boundary conditions.

The quality metrics for the individual examples have been compiled in Table \ref{tab:2Daccuracies}. Furthermore, we show in Table \ref{tab:predictiontime} the time required by our transfer learning-based method to make predictions for several 2D optimal topologies: the average prediction time is 0.017 seconds per design case versus 138.0 seconds for the SIMP solver. 

These resulting quality metrics are very promising, and are particularly so for the resolutions considered in the our experiments. Moreover, as the resolution increases the level of detail that is picked up by our transfer learning-based predictor increases as well. Notably, the normalized prediction time increases at a much slower rate compared to that of the SIMP algorithm.

\begin{table*}[h!]
    \centering
    \caption{2D structures: MSE, binary accuracies, and compliance error relative to SIMP.}
    \begin{tabular}{lcSSSSS}
        \toprule
        Design Domain    & Resolution & {\shortstack{Number of \\ test cases}} & {MSE}  & {\shortstack{Binary \\Accuracy}} & {\shortstack{Compliance \\ Error}} & {\shortstack{Compliance \\ Error Std.}}\\
        \midrule
        from Fig. \ref{fig:2Dbeam-recta} (predicted by the source network)  & 40 x 80 & 2000 & 2.14\% & 96.61\% & 2.1\% & 0.054     \\
        from Fig. \ref{fig:2Dbeam-rectb}  & 80 x 160  & 750  & 3.70\%  & 94.50\% & 3.65\% & 0.055  \\
        from Figure \ref{fig:2Dbeam-rectc}  & 120 x 160 & 500  & 3.45\%  & 94.61\% & 4.81\% & 0.056  \\
        from Figure \ref{fig:2Dbeam-rectd}  & 120 x 240 & 625  & 4.83\%  & 94.59\% & 4.93\% & 0.062\\
        from Figure \ref{fig:2Dbeam-recte}  & 160 x 320 & 375  & 4.29\%  & 94.46\% & 6.54\% & 0.080    \\
        from Figure \ref{fig:2Dbeam-rectf}  & 200 x 400 & 375  & 4.35\%  & 94.55\% & 9.57\% & 0.106  \\
        Curved beam (Fig. \ref{fig:9a}) & 80 x 160 & 500  & 4.01\%  & 94.44\% & 9.2\% & 0.088\\
        Curved beam (Fig. \ref{fig:9a}) & 120 x 240 & 750  & 4.13\%  & 93.94\% & 11.3\% & 0.091 \\
        Frame (Fig. \ref{fig:9b}) & 80 x 160 & 500  & 2.61\%  & 95.53\% & 5.39\% & 0.070 \\
        Frame (vol. fr.  = 0.4) (Fig. \ref{fig:9b}) & 120 x 240 & 500  & 3.00\%  & 94.80\% & 15.50\% & 0.182 \\
        L shape (vol. fr. = 0.3) (Fig. \ref{fig:7a}) & 120 x 240 & 500  & 2.50\%  & 95.75\% & 13.20\% & 0.130 \\
        L shape w/hole (vol. fr. = 0.3) (Fig. \ref{fig:7b}) & 120 x 240 & 500  & 2.60\%  & 95.72\% & 20.00\% & 0.210 \\
  		\bottomrule
  		 		\textit{Average} &	&	&  3.46\% & 94.95\% & 8.85\% & 0.098 \\
        \bottomrule
    \end{tabular}
    
    \label{tab:2Daccuracies}
\end{table*} 
 
 \begin{table}[h!]
    \centering
    \caption{2D Structures: Comparison of computational time of our predictions vs. the SIMP algorithm.}
    \begin{tabular}{l S S}
        \toprule
        Resolution & {\shortstack{SIMP \\(sec. per case)}} & {\shortstack{Our method \\(sec. per case)}} \\
        \midrule
        80 x 160 (Fig. \ref{fig:2Dbeam-rectb}) & 24   & 0.0093      \\
        120 x 160 (Fig. \ref{fig:2Dbeam-rectc}) & 36   & 0.010      \\
        120 x 240 (Fig. \ref{fig:2Dbeam-rectd}) & 80   & 0.015       \\
        160 x 320 (Fig. \ref{fig:2Dbeam-recte}) & 200   & 0.022       \\
        200 x 400 (Fig. \ref{fig:2Dbeam-rectf}) & 350  & 0.030      \\
        \bottomrule
         \textit{Average}	& 138 & 0.017		\\
        \bottomrule
    \end{tabular}
    \label{tab:predictiontime}
\end{table}

\begin{figure}[h!]
    \centering
    \includegraphics[width=\Columnwidth]{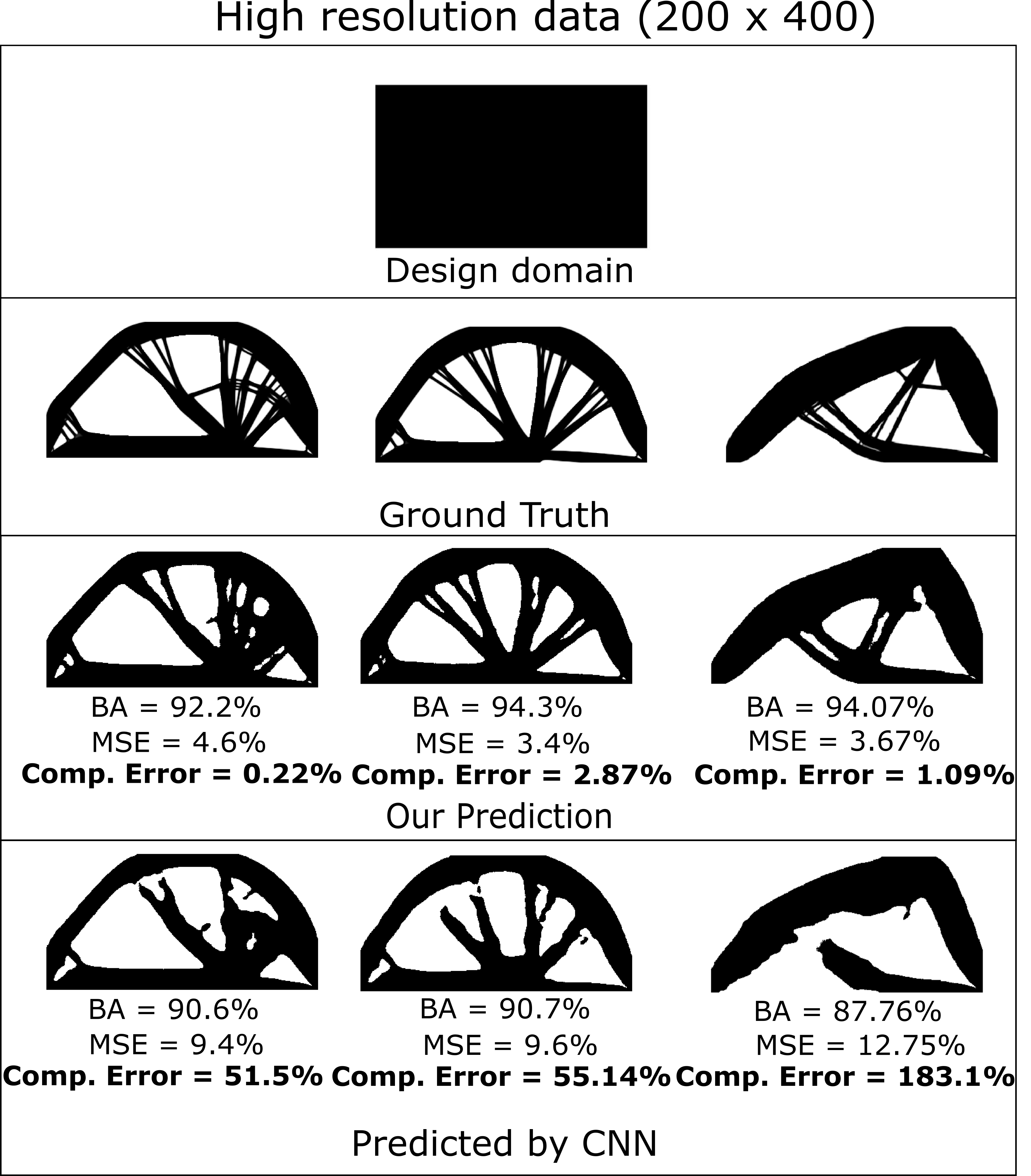}
    \caption{Comparison between our predictions and those of a traditional CNN-based algorithm trained on 1500 and 1650 high resolution data, respectively. Our source network has been trained with low resolution cases as well, and we took into account the time required to generate these low resolution cases when selecting the number of high resolution cases for the traditional CNN-based algorithm.}
    \label{fig:8}
\end{figure}

\subsubsection*{Comparison With Traditional Deep Learning Methods}
 
In order to compare our transfer learning-based method with other published  deep learning methods for topology optimization, we examined two criteria. Specifically, we compare our method in terms of:
 \begin{enumerate}
     \item time required to generate equivalent training data producing similar prediction performance; and
     \item prediction accuracy with the training data generated in the same amount of time.
 \end{enumerate}
 
One key advantage of our method compared to the traditional deep learning methods is that the method proposed in this paper requires a much smaller number of high-resolution cases to train the target network than to train an equivalent deep learning network. For example, we used 8,000 low resolution 2D cases and 1500 high resolution cases to obtain the high quality predictions shown in Figures \ref{fig:2Dbeam-rect}-\ref{fig:9}. On top of that, the source network has to be trained only once with the 8,000 low resolution 2D cases.

On the other hand, training a deep learning network to produce a prediction of similar quality requires at least 8,000 high resolution cases for every design space and every type of boundary conditions. This is a significant time difference that becomes more severe as the resolution of the design space increases and as we move to 3D. For example, the SIMP algorithm  \cite{andreassen2011efficient} requires 5 seconds to produce the optimal topology for every low resolution (40 x 80) case and 350 seconds for every high resolution case (200 x 400). Thus, training our source and target networks requires \textit{5.3 times} fewer high resolution cases than the equivalent CNN, and about $5$ times less computation time. To put this in perspective, by using  this particular SIMP algorithm,  generating equivalent training datasets for our method is $~620$ hours faster than for an equivalent CNN. This difference rapidly increases with the increase in the fidelity of the desired results.

To be able to compare the prediction accuracies of transfer learning based vs deep learning based methods for the same amount of time, we replicated our target network in terms of layers, loss function, optimizer and so on, and trained it as a normal deep learning network. We determined the average total time needed to generate the training sets for our method for the example shown in Figure \ref{fig:2Dbeam-rectf}, and then generated as many high-resolution cases for the deep CNN as possible in the same amount of time. Finally, we trained the deep CNN network with this dataset and compared the quality of the predictions between our method and of the corresponding deep CNN, as illustrated in Figure \ref{fig:8}.  This experiment clearly, although unsurprisingly,  illustrates the significant superiority of our predictions compared to other deep learning methods based on data generated in the \textit{same} amount of time.
 
Furthermore, the much smaller size of the training dataset required by our transfer learning network implies that the corresponding training time is also much smaller than that required by traditional deep learning networks, as illustrated in Table \ref{tab:2Dtrainingtime}.

  \begin{table*}[h!]
    \centering
    \caption{2D Structures: Comparison of the corresponding training time. The last two rows show the training time required by an equivalent Deep CNN.}
            \sisetup{group-separator={,},group-minimum-digits = 4}
    \begin{tabular}{ l S S S S}
        \toprule
       {Resolution} & {\shortstack{Number of \\ training cases}} & {\shortstack{Training time  \\ (seconds, per epoch)}} & {\shortstack{Number of \\ epochs}} & {\shortstack{Training time \\  (minutes)}}  \\
        \midrule
        40 x 80 (Fig. \ref{fig:2Dbeam-recta}) & 8000 & 62.87 & 29 & 30.3        \\
        80 x 160 (Fig. \ref{fig:2Dbeam-rectb}) & 1500 & 26 & 4 & 1.73 \\
        120 x 160 (Fig. \ref{fig:2Dbeam-rectc}) & 1500 &27.19 & 5 & 2.26      \\
        120 x 240 (Fig. \ref{fig:2Dbeam-rectd}) & 1500 & 34.88 &10 & 5.81     \\
        160 x 320 (Fig. \ref{fig:2Dbeam-recte}) & 1500 & 41.25 &8 & 5.5   \\
        200 x 400 (Fig. \ref{fig:2Dbeam-rectf}) & 1500 & 58.68 &9 & 8.80  \\
        \bottomrule
        200 x 400 (CNN) (Fig. \ref{fig:8}) & 1650 & 67.43 &28 & 31.46 \\
        200 x 400 (CNN) & 8000 & 307.54 & 29 & 148.64 \\
        \bottomrule
    \end{tabular}
    \label{tab:2Dtrainingtime}
\end{table*}

\subsection{3D structures} \label{3DStructures}

\begin{figure*}[h!]
\vspace{3pt}
\begin{center}
    \begin{tabular}{cc}
        \includegraphics[width=\Columnwidth]{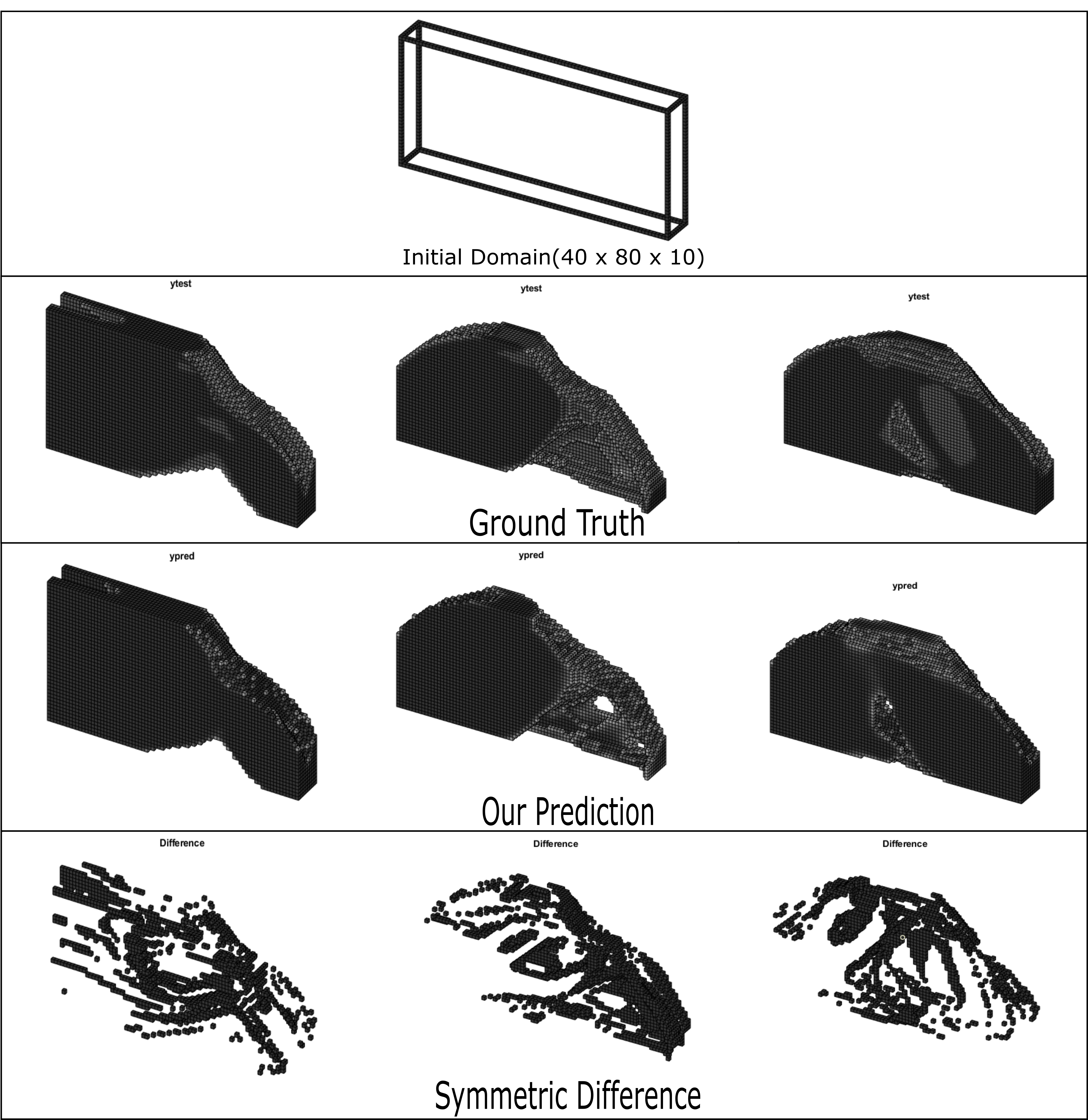}
        \hspace{15pt}
        &
        \includegraphics[width=\Columnwidth]{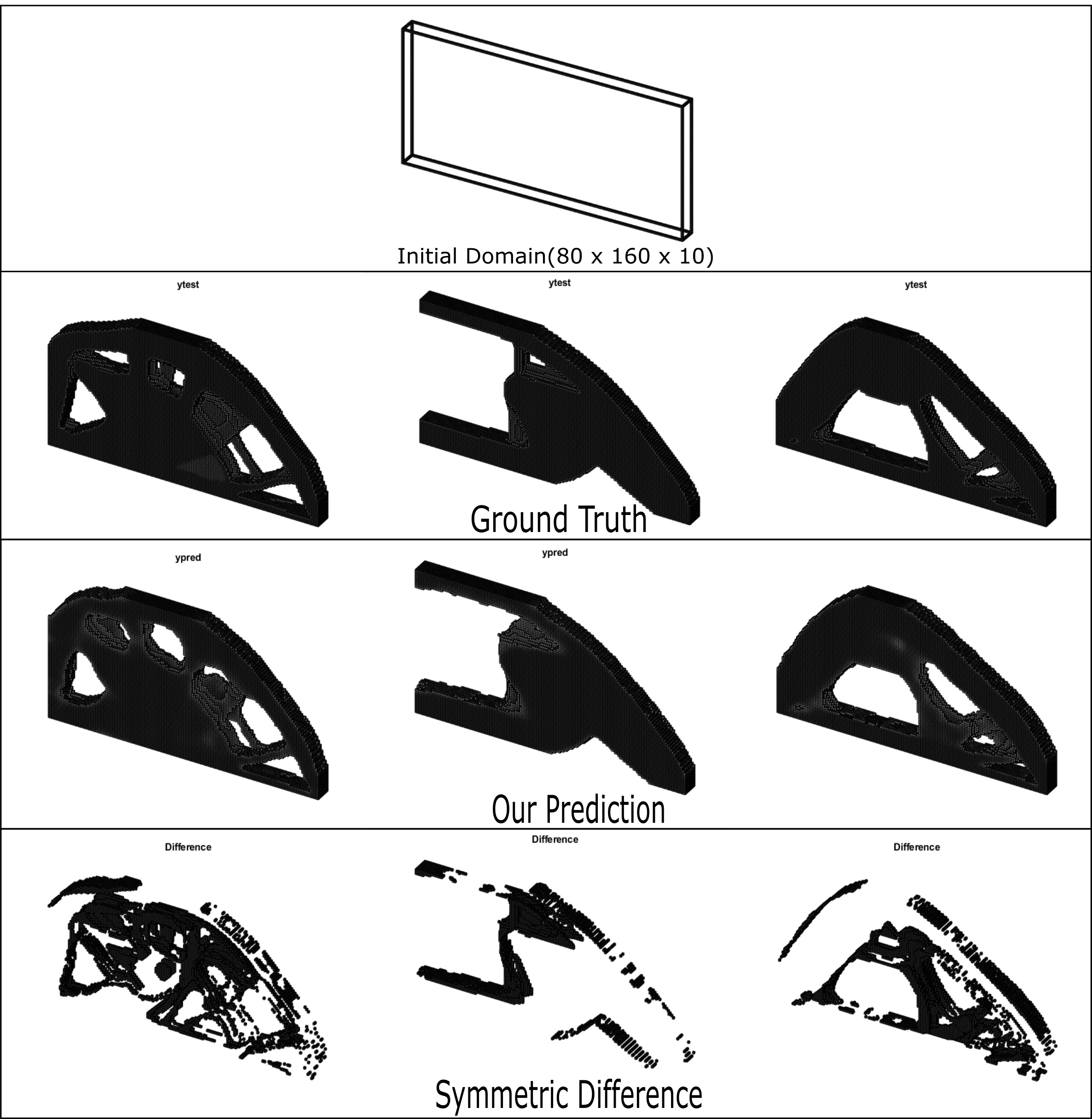}
        \hspace{15pt}
        \\
        (a) & (b)\\
    \end{tabular}
    \\
    \begin{tabular}{cc}
        \includegraphics[width=1\Columnwidth]{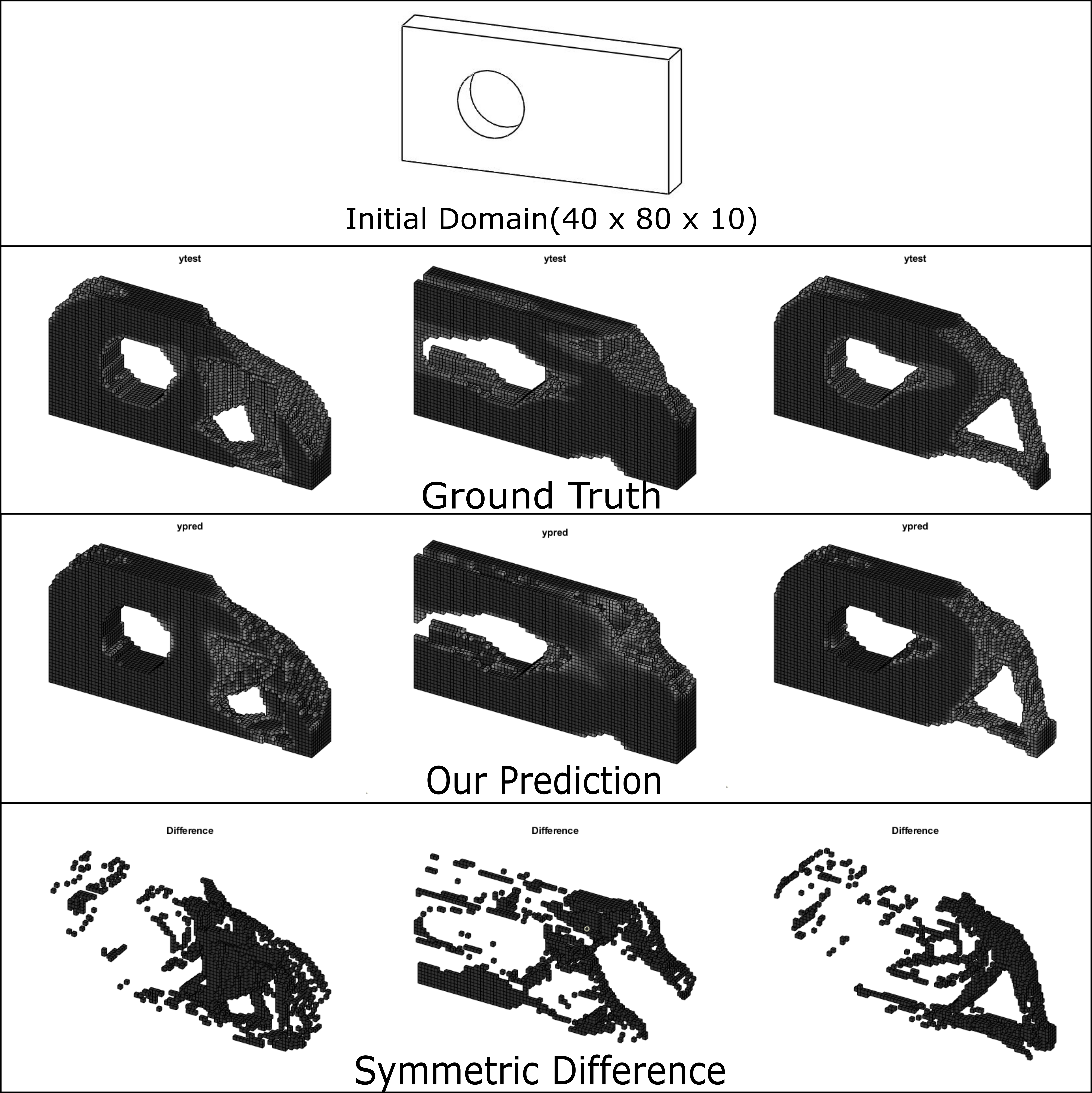}
        \hspace{15pt}
        &
        \includegraphics[width=1\Columnwidth]{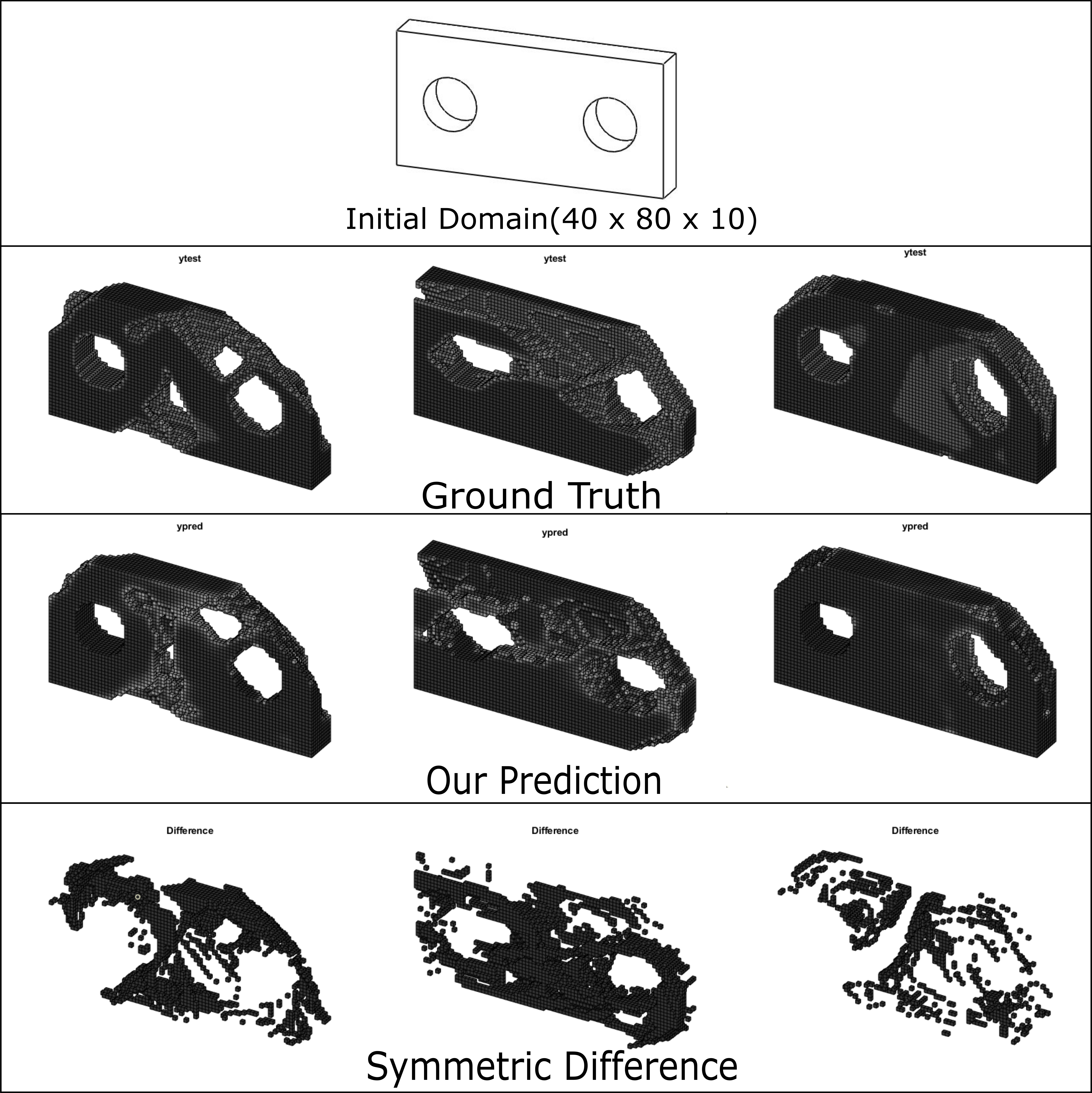}
        \hspace{15pt}
        \\
        (c) & (d) \\
    \end{tabular}
    \\
\end{center}
\caption{The predicted and ground truth optimal structures for 3D design spaces, including  their symmetric difference. Figures (a) and (b) show a simple parallelepipedic domain with two resolutions and figures (c) and (d) show a beam with one or two holes, respectively. The individual quality metrics for our predictions are presented in Table \ref{tab:3Daccuracies}, and the prediction time is shown in Table \ref{tab:3Dpredictiontime}.} \label{fig:3Dbeam} \vspace{3pt} \hrule
\end{figure*}

\begin{figure*}[h!]
 	\centering
	\begin{subfigure}[t]{\textwidth}
 		\centering
 		\includegraphics[width=2\Columnwidth]{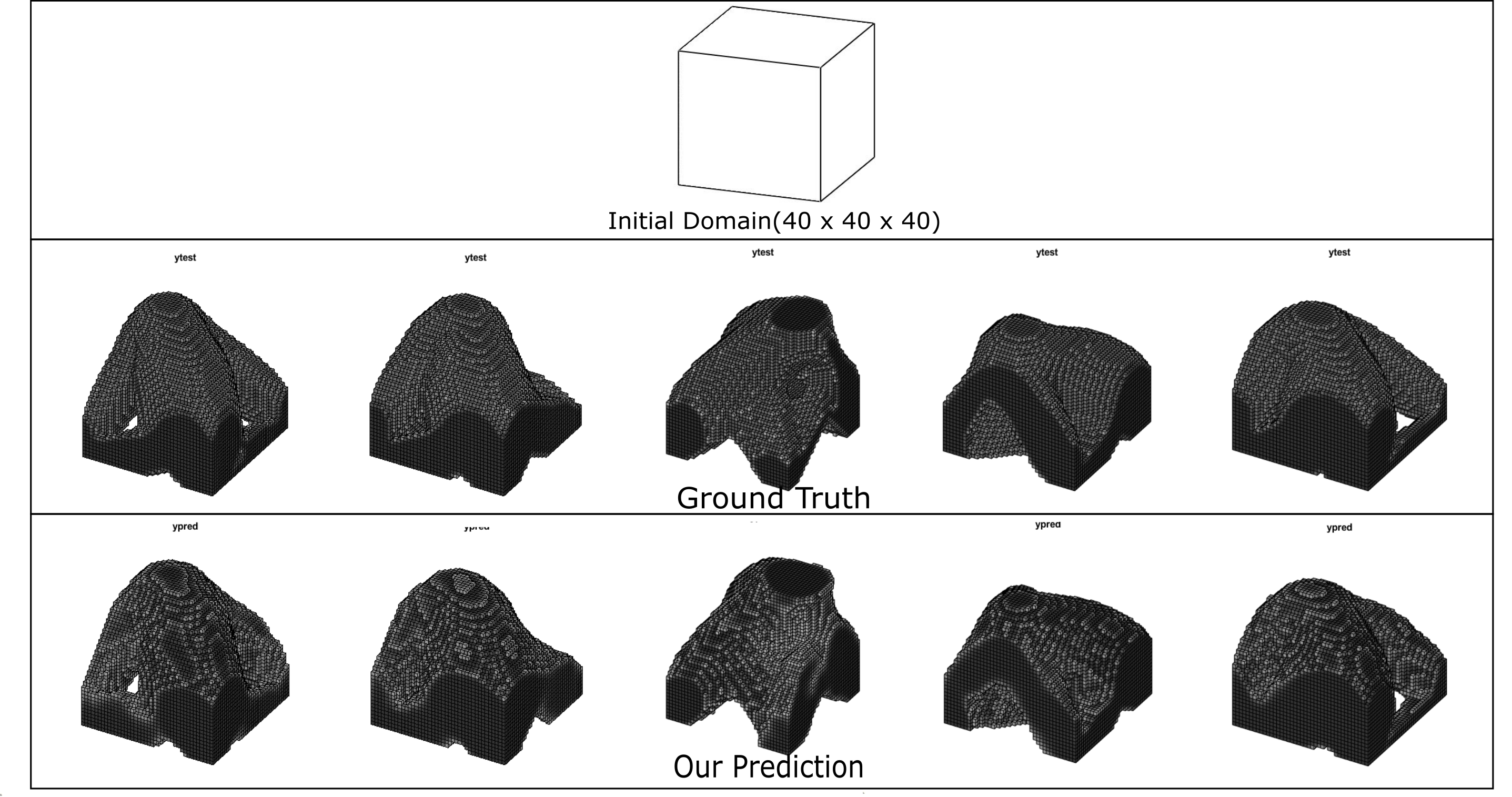}
 		\caption{}
 		\label{fig:cubea}
 	\end{subfigure}
 	\hspace{-0.3cm}
 	\begin{subfigure}[t]{\textwidth}
 		\centering
 		\includegraphics[width=2\Columnwidth]{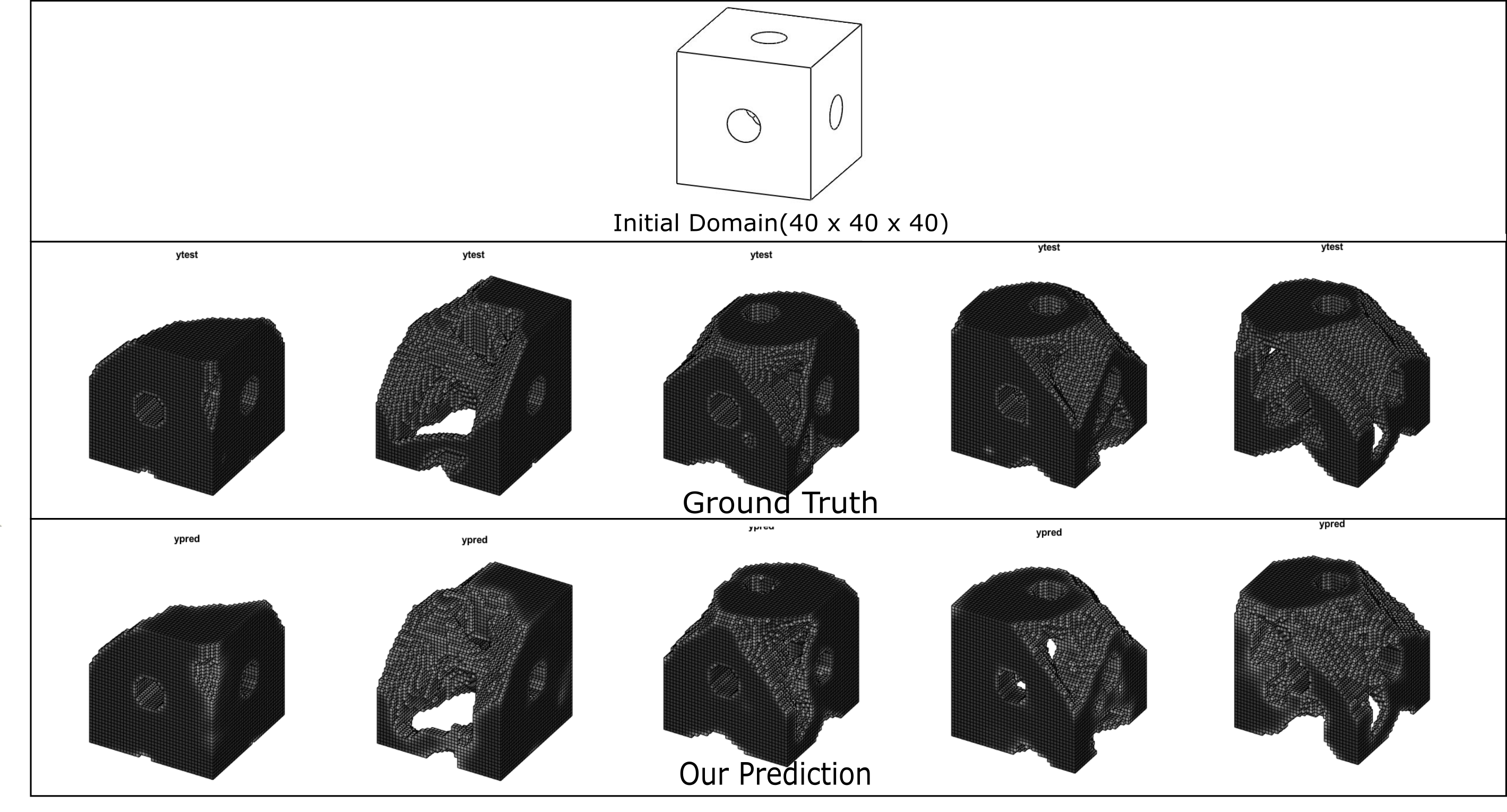}
 		\caption{}
		\label{fig:cubeb}
 	\end{subfigure} 
 	\caption{Predicted versus ground truth structures for two different domains as well as the boundary conditions shown in Figure \ref{fig:3}(d). Both domains and  boundary conditions are unseen to the source network. The individual quality metrics for our predictions are presented in Table \ref{tab:3Daccuracies}, and the prediction time is shown in Table \ref{tab:3Dpredictiontime}.}
 	\label{fig:cube}
 \end{figure*} 
 
 \begin{figure*}[h!]
	\centering
	\includegraphics[width=\textwidth]{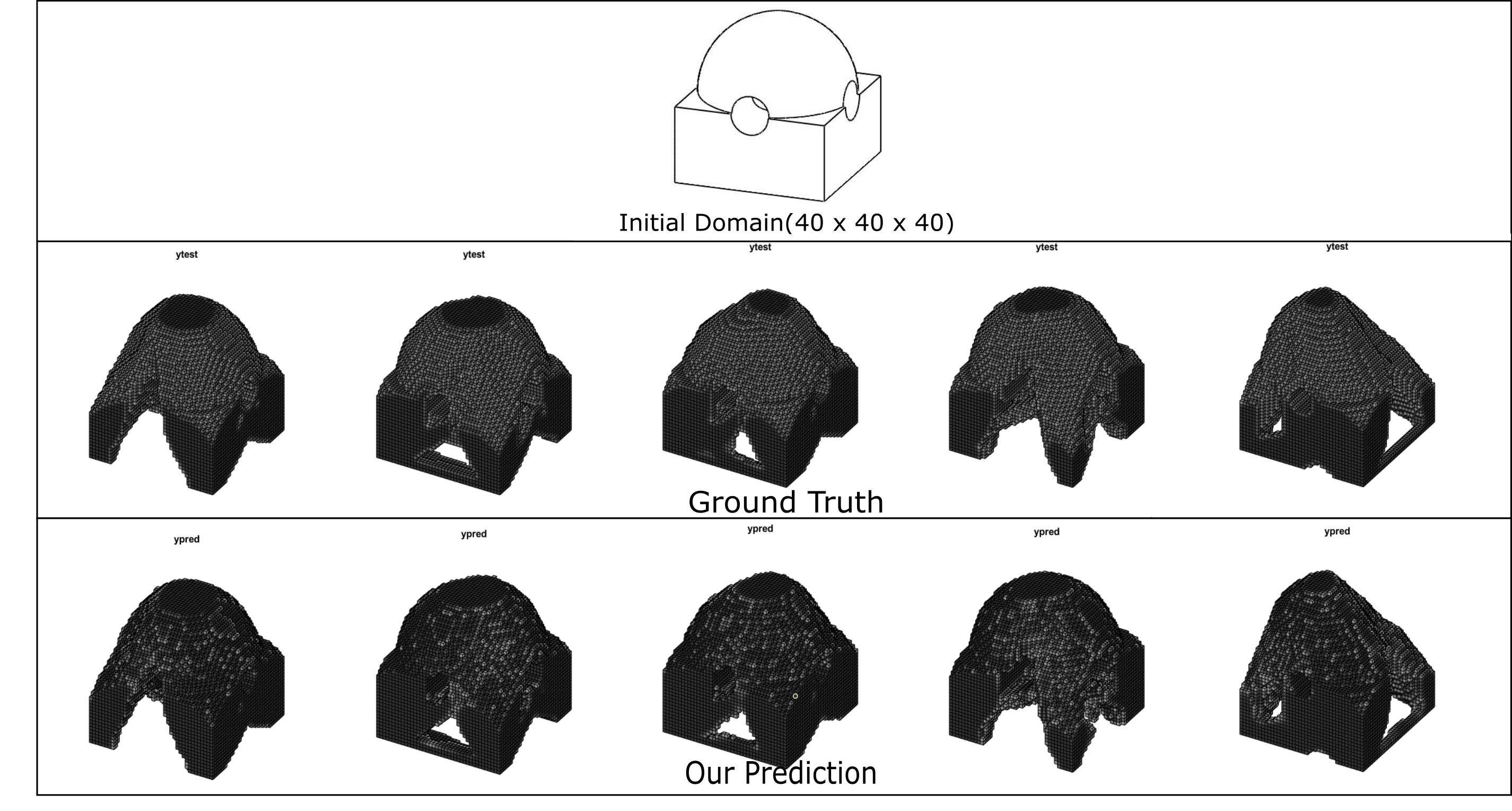}
	\caption{A domain and boundary conditions that were not in the training set for our source model, but it was included in the much smaller dataset used for fine-tuning the target model. The individual quality metrics for our predictions are presented in Table \ref{tab:3Daccuracies}, and the training time is shown in Table \ref{tab:3Dtrainingtime}.}
	\label{fig:cubesph}
\end{figure*}

We also applied our transfer-learning based method to 3D domains and we used a freely available Matlab Code \cite{liu2014efficient} to generate the ground truth results. For the 3D examples, we used 12,000 low resolution data ($20 \times 40 \times 10$) to train the source model, and 1500 high(er) resolution data for fine-tuning the target model. Solving the TO problem in 3D is notoriously time consuming. Thus,  to generate the ground truth cases for the examples used in this section, we limited the number of iterations of the SIMP solver to 150. To generate the training datasets for all 3D examples with parallelepipedic domains (Figures \ref{fig:3Dbeam}), the boundary conditions were randomly chosen from one of the 3 cases shown in Figures \ref{fig:3}(a-c), and the location, orientation and magnitude have also been randomized. The cases shown in Figures \ref{fig:cube} and \ref{fig:cubesph} used randomized boundary conditions according to Figure\ref{fig:3}(d).

We first show in Figures \ref{fig:3Dbeam}(a) \& (b) the comparison between our predicted and the ground truth (SIMP) 3D optimal structures for two beams obtained for two different resolutions of the design space, namely $40 \times 80 \times 10$ and $80 \times 160 \times 10$. Moreover, Figures \ref{fig:3Dbeam}(c) \& (d) show the same comparison for beams that have different topologies for a $40 \times 80 \times 10$ resolution. In order to help illustrate the difference between the two solutions more clearly,  we also provide the symmetric difference between the ground truth and predicted result, i.e., the voxels that are in either one but not the other structure.

Importantly, our method can provide impressive performance even for cases for which the source network has not been specifically trained for. As an example, consider the design space illustrated in Figure \ref{fig:cubesph}, which was not part of the training set for our source model, but it was included in the much smaller dataset used to fine-tune the target model. The predicted optimal structures for this new design problem, which are summarized in Figures \ref{fig:cube} \& \ref{fig:cubesph} are still highly accurate. Table \ref{tab:3Daccuracies} shows the average MSE, binary accuracy and compliance error for the 3D cases described above. Note that average MSE and binary accuracy are also around 3\% and 95\% for these 3D cases.  Furthermore, we summarize in Table \ref{tab:3Dpredictiontime} the  time required by our method to predict the 3D optimal structures, and compare these times with those required by the SIMP method to reach 150 iterations for the same 3D problems. Our method is consistently multiple orders of magnitude faster than the SIMP method, and achieves real-time rates even for our preliminary and non-optimized implementation.
 
 \begin{table*}[ht!]
    \centering
    \caption{3D structures: MSE, Binary Accuracy and Compliance Error relative to SIMP.}
    \begin{tabular}{lcSSSS}
        \toprule
        Design Domain    & Resolution & {\shortstack{Number of \\ test cases}} & {MSE}  & {\shortstack{Binary \\Accuracy}} & {\shortstack{Compliance \\ Error}}\\
        \midrule
        Domain  & 20 x 40 x 10 & 1000 & 2.04\% & 95.62\% & 1.56\%    \\
        Domain (Fig. \ref{fig:3Dbeam}a)  & 40 x 80 x 10 & 300 & 3.14\% & 94.31\% & 2.43\%    \\
        Domain (Fig. \ref{fig:3Dbeam}b)  & 80 x 160 x 10 & 100 & 3.1\% & 93.9\% & 10.1\%     \\
        With hole (Fig. \ref{fig:3Dbeam}c)  & 40 x 80 x 10 & 150 & 3.45\% & 94.00\% & 2.05\%    \\
        With 2 hole (Fig. \ref{fig:3Dbeam}d)  & 40 x 80 x 10 & 150 & 3.52\% & 94.31\% & 2.85\%  \\
        Cube (Fig. \ref{fig:cubea})  & 40 x 40 x 40 & 175 & 3.28\% & 93.29\% & 9.9\% \\
        Cube with holes (Fig. \ref{fig:cubeb})  & 40 x 40 x 40 & 180 & 3.51\% & 93.11\% & 7.5\%  \\
        Dome with holes (Fig. \ref{fig:cubesph})  & 40 x 40 x 40 & 200 & 2.41\% & 95.71\% & 0.38\%  \\
  		\bottomrule
  		 		\textit{Average} &	&	&  3.05\% & 94.28\% & 4.60\%  \\
        \bottomrule
    \end{tabular}
    
    \label{tab:3Daccuracies}
\end{table*} 
 
        
    

\begin{table}[ht!]
    \centering
    \caption{3D structures: comparison of prediction time vs. SIMP algorithm.}
    \sisetup{group-separator={,},group-minimum-digits = 4}
    \begin{tabular}{lSS}
        \toprule
        Resolution & {\shortstack{SIMP \\(sec. per case)}} & {\shortstack{Our method \\(sec. per case)}} \\
        \midrule
        20 x 40 x 10 & 300  & 0.015     \\
        40 x 80 x 10 (Fig. \ref{fig:3Dbeam}a) & 4500  & 0.031     \\
        80 x 160 x 10 (Fig. \ref{fig:3Dbeam}b) & 7500  & 0.04     \\
        40 x 40 x 40 (Fig. \ref{fig:cube}) & 5550  & 0.033     \\
        \bottomrule
        \textit{Average}	&  4462.5 & 0.029 \\
        \bottomrule
    \end{tabular}
    \label{tab:3Dpredictiontime}
\end{table} 
 
%

\begin{table*}[ht!]
    \centering
    \caption{3D structures: Training time}
            \sisetup{group-separator={,},group-minimum-digits = 4}
    \begin{tabular}{lSSSS}
        \hline
               {Resolution} & {\shortstack{Number of \\ training cases}} & {\shortstack{Training time  \\ (seconds, per epoch)}} & {\shortstack{Number of \\ epochs}} & {\shortstack{Training time \\  (minutes)}}  \\
        \hline
        20 x 40 x 10 & 12000 & 1101.87 & 20 & 367.25 \\
        40 x 80 x 10 (Fig. \ref{fig:3Dbeam}a) & 1500 & 174.56  & 5 & 14.54      \\
        80 x 160 x 10 (Fig. \ref{fig:3Dbeam}b) & 1500 & 290  & 15 & 72.5      \\
        40 x 40 x 40 (Cube) (Fig. \ref{fig:cubea}) & 1900 & 252  & 20 & 84      \\
        40 x 40 x 40 (Cube with holes) (Fig. \ref{fig:cubeb}) & 1700 & 230  & 30 & 115      \\
        40 x 40 x 40 (Dome with 2 holes) (Fig. \ref{fig:cubesph}) & 1500 & 202  & 35 & 117.8      \\
        \hline
    \end{tabular}
    \label{tab:3Dtrainingtime}
\end{table*} 

Similarly with the 2D case described above, the time required to generate a comparable training dataset requires much less time for our method  relative to the time required for an equivalent deep CNN. By extrapolating from our 3D experiments, our method requires a time that is at least 5 times less than the time required to generate the comparable dataset for the equivalent deep CNN, and the difference increases with the resolution. For example, for the case shown in Figure \ref{fig:cubesph}, by using the algorithm described in \cite{liu2014efficient} run in parallel on 100 processors, for a $40 \times 40 \times 40$ resolution, and with the optimization stopped early at the $150^{th}$ iteration for each case, the dataset generation for our method requires approximately 152 hours (6.3 days) less than the time required to generate an equivalent dataset for the deep CNN, even with such an aggressive parallelization. 
At higher resolutions, and assuming that one can use a highly parallelized algorithm, such as the one implemented on the GPU described in \cite{wu2015system} that computes an optimal topology for a cantilever beam for a $200 \times 100 \times 100$ resolution in 144 seconds, our transfer learning-based method would need at least 388 hours (16.2 days!) less time than the time required to generate the equivalent training set for the deep CNN.


\subsection{Generalizability of Network Predictions}

The solution manifolds between our low-resolution and high-resolution domains for some of the examples shown in this paper can be considered to already be dissimilar. Specifically, our source network is trained on rectangular/parallelepipedic  low-resolution domains of genus 0, but the target network is fine-tuned on domains whose geometry, topology and boundary conditions are unseen to the source network. Nevertheless, to further confirm the generalization performance of our method, we also considered the types of problems from \cite{cang2019one}, which use local density constraints for the topology optimization. It is important to note that we used here the same source network trained on the dataset as explained above, but only fine-tuned the target network on the data set used in \cite{cang2019one}. 

\begin{figure*}[ht!]
\centering
    \begin{subfigure}[t]{.5\textwidth}
    \centering
      \includegraphics[width=\Columnwidth]{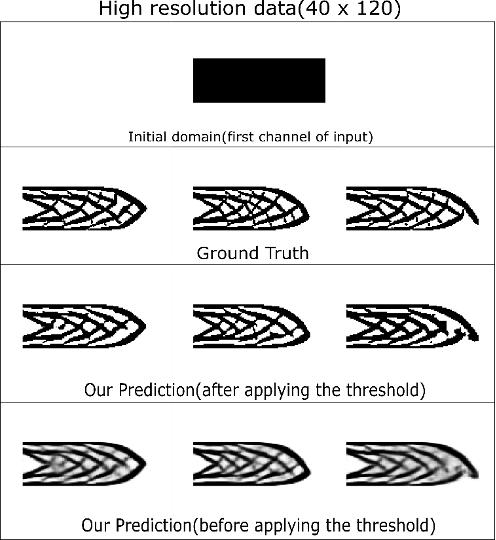}
      \caption{}
    \end{subfigure}
    \hspace{-0.3cm}
    \begin{subfigure}[t]{.5\textwidth}
    \centering
      \includegraphics[width=\Columnwidth]{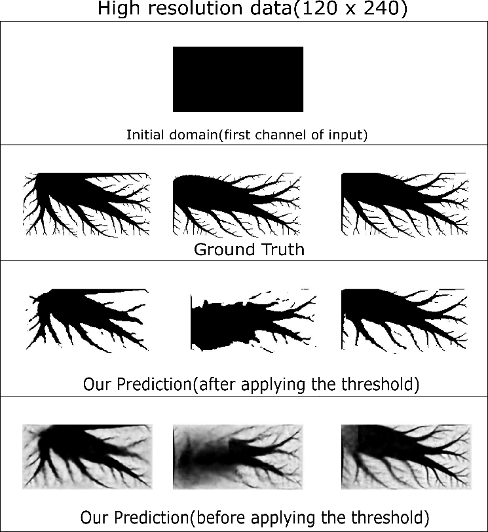}
      \caption{}
    \end{subfigure} 
 \caption{Predicted optimal structures versus ground truth (SIMP optimized) for problems with local density constraints \cite{cang2019one} and heat sink design. The fact that we used the same source network for these new, more complex TO problems strongly affirms that knowledge is being transferred between our source and target  networks.}
\label{fig:generalizability}
\end{figure*}

The results are shown in Figure \ref{fig:generalizability}(a). Note that the predictions of our network are visually very similar to the ground truth. Moreover, the work presented in \cite{cang2019one} used 7000 training cases, while our target network only required 1500 cases for fine-tuning. The fact that we used the same source network for this new, more complex TO problem strongly affirms that knowledge is being transferred between our source and target  networks. 

Moreover, we have also explored the heat sink design problem, which results in a tree-like optimal structure with very thin branches. The ground truth solution in this case has genus 0, so this can be considered more of a size rather than a topology optimization problem.

We adapted our network to the heat sink design problem to further illustrate the capabilities of our method and of the knowledge transfer abilities between our source and target network.  Figure \ref{fig:generalizability}(b) shows the network performance for some of these test cases, including the predicted optimal solutions before and after the application of the density threshold, i.e., integer rounding for the density values. While capturing the very thin members of the optimal structure remains a limitation, which is discussed in the next section, our results indicate that knowledge is transferred between the source and the target network, as intended, even for the case of heat sink design.

It is also important to note that in all the examples presented in the paper we used the same source network trained only once, as described above.

\subsection{Current Limitations in Capturing Thin Members}

Our transfer learning network has not been designed to achieve high prediction performance of very thin members. We hypothesize that there are several reasons for this behavior. First, our current network does not have a complexity that is sufficiently high, mainly due to GPU limitations imposed by our hardware on the size of the layer output during training. A more complex network would definitely improve detection performance for the thin members. On the other hand, a more complex network would need more training data, which is expensive to generate with the SIMP algorithms that we used. Second, the thin members that we predict with a somewhat lower accuracy have a width of 1-2 pixels/voxels, which is at or below 1\% of the largest dimension of the domain. Increasing the resolution relative to the width of the members would also improve prediction performance for the thin members. This, in turn, would require higher resolution training data, whose computational cost increases exponentially with the resolution, as well as a more complex network architecture. Finally, we applied in all our examples a simple integer rounding as a threshold for the densities of the individual elements. This, in turn, eliminated from the solutions that we report all pixels/voxels whose density was below 0.5, including those that belonged to the thin members. However, our network predicts a much more nuanced density field, as shown in the last row of Figure \ref{fig:generalizability}.

\subsection{Transfer Learning in Conjunction with SIMP}

One way to take full advantage of the information output by our network without increasing the complexity of the network architecture and the data resolution is to couple our transfer learning-based predictions to SIMP to obtain a more accurate definition of the thin members. Table \ref{tab:TO+TL}  shows the time required for the SIMP algorithm to generate the optimal structure by using as input the prediction output by our network. Specifically, for a 200 x 400 resolution domain, the SIMP algorithm that we used calculates such an optimum structure in 8 seconds (with an average of 4.5 seconds) compared to 350 seconds needed by a normal SIMP optimization starting with the full design domain.

\begin{table}[ht!]
    \centering
    \caption{Average time required by SIMP to generate the optimal solution starting from our predicted structure.}
    \begin{tabular}{l S}
        \toprule
        Resolution & {\shortstack{Time \\(sec. per case)}}\\
        \midrule
        80 x 160 (Fig. \ref{fig:2Dbeam-rectb}) & 1       \\
        120 x 160 (Fig. \ref{fig:2Dbeam-rectc}) & 3.5       \\
        120 x 240 (Fig. \ref{fig:2Dbeam-rectd}) & 4         \\
        160 x 320 (Fig. \ref{fig:2Dbeam-recte}) & 6        \\
        200 x 400 (Fig. \ref{fig:2Dbeam-rectf}) & 8      \\
        \bottomrule
         \textit{Average}	& 4.5 		\\
        \bottomrule
    \end{tabular}
    \label{tab:TO+TL}
\end{table} 

\section{Conclusions} \label{conclusions:sec}
 
We proposed in this paper a highly efficient and accurate non-iterative topology optimization method that uses transfer learning on a convolutional neural network architecture. Our method uses low resolution datasets to train a source network and a much smaller high resolution dataset to fine-tune a target network. The learned knowledge captured by the source network once is transferred to the target network, so that the latter requires a much smaller number of training cases than an equivalent deep CNN to make predictions with the same level of accuracy. We provided numerous examples to show that the proposed method produces predictions of the optimal 3D topologies at real-time rates for non-trivial 3D high resolution TO problems. Furthermore, we showed that the proposed method can produce accurate predictions efficiently for various design spaces, boundary conditions, and volume fractions, including for cases that have not been part of the source network's training set. Our experiments achieved an average binary accuracy and MSE around 95\% and 3\%, respectively, at real-time rates in both 2D and 3D.

Like any other data-driven method, our approach inherits any existing data biases in the datasets. In our experiments, we reduced the biases by randomizing the input used to generate the ground truth structures. Moreover, the capability of this method to explore regions of the design space for which the algorithm has not been trained for has the same limitations as most other transfer learning algorithms. In addition, when the source task and the target task are not similar enough, negative transfer may occur and the algorithm performance may fail to improve \cite{torrey2010transfer} without additional information. This, however, is not unlike what happens in real life. Consider one of the traditional examples used to explain transfer learning, namely that of learning how to ride a bicycle. Clearly these skills can be transferred by bicycle riders and used to learn how to ride other two wheeled devices, such as motorcycles or scooters. However, the same bicycle riding skills cannot be easily employed to ride, for example, unicycles - as anyone that has tried to ride a unicycle can attest to.

Perhaps the key bottleneck of any data driven TO method is the computational cost to generate suitable training data, which is computationally demanding for gradient-based optimization algorithms, and is particularly so in 3D. On one hand, our experiments show that the proposed transfer learning-based method requires much less time than equivalent deep CNN to generate the training dataset to reach the same accuracy. On the other hand,  employing more efficient gradient-based approaches to generate ground truth optimal structures are needed to be able to perform careful studies of how to best train the proposed  transfer learning-based method and to better understand its generalization capabilities, scalability and limitations. Fortunately, recent advances in software and hardware architectures, such as the recently announced optimized physics libraries from AMD and NVIDIA, which promise to include FEA capabilities, come at the right time and with the potential to dramatically speed up the data generation for our purposes. 

Generalizability is a critical aspect of any machine learning-based method. On one hand, we discussed In section \ref{results} the capability of the proposed transfer learning method to transfer knowledge between the source and the target network, and we illustrated the generalizability of its predictions. However, a more complete treatment of this difficult question, which has not been addressed yet in the literature, would likely involve  constructions of local subspaces that approximate the solution manifolds corresponding to the low and high resolution problems, and definitions of new metrics that would measure the ``distance'' between the corresponding approximations of the solution manifolds. Very likely, an effective distance would probably require the projection of these approximations of the solution manifolds onto some common subspace, but there are a number of important open problems that need to be solved first. Perhaps the current efforts in the mathematical optimization community focused on Reduced Bases Methods could provide some insight into these issues.

It is also possible that modifying the type of input to replace or augment the explicit boundary conditions by one or more physical fields, such as displacement and stress/strain, would further improve the generalization capabilities of the proposed method. However, such a study is outside the scope of this paper.

Nevertheless, the proposed approach shows that transfer learning can serve as a practical underlying framework for performing real-time 3D design space  explorations with topology optimization. To the best of our knowledge, this paper documents the first attempt to use transfer learning for topology optimization and provides exciting and important directions for future research.  

\section*{Acknowledgement}
This work was supported in part by the National Science Foundation grants CMMI-1462759, IIS-1526249, and CMMI-1635103.





\bibliography{main}

\begin{thebibliography}{10}
\expandafter\ifx\csname url\endcsname\relax
  \def\url#1{\texttt{#1}}\fi
\expandafter\ifx\csname urlprefix\endcsname\relax\def\urlprefix{URL }\fi
\expandafter\ifx\csname href\endcsname\relax
  \def\href#1#2{#2} \def\path#1{#1}\fi

\bibitem{sigmund2013topology}
O.~Sigmund, K.~Maute, Topology optimization approaches, Structural and
  {M}ultidisciplinary {O}ptimization: {A} {C}omparative {R}eview 48~(6) (2013)
  1031--1055.

\bibitem{amir2011reducing}
O.~Amir, O.~Sigmund, On reducing computational effort in topology optimization:
  how far can we go?, Structural and {M}ultidisciplinary {O}ptimization 44~(1)
  (2011) 25--29.

\bibitem{limkilde2018reducing}
A.~Limkilde, A.~Evgrafov, J.~Gravesen, On reducing computational effort in
  topology optimization: we can go at least this far!, Structural and
  {M}ultidisciplinary {O}ptimization 58~(6) (2018) 2481--2492.

\bibitem{banga20183d}
S.~Banga, H.~Gehani, S.~Bhilare, S.~Patel, L.~Kara, {3D} topology optimization
  using convolutional neural networks, arXiv preprint arXiv:1808.07440 (2018).

\bibitem{li2019non}
B.~Li, C.~Huang, X.~Li, S.~Zheng, J.~Hong, Non-iterative structural topology
  optimization using deep learning, Computer-{A}ided {D}esign (2019).

\bibitem{yu2019deep}
Y.~Yu, T.~Hur, J.~Jung, I.~G. Jang, Deep learning for determining a
  near-optimal topological design without any iteration, Structural and
  {M}ultidisciplinary {O}ptimization 59~(3) (2019) 787--799.

\bibitem{bendsoe1989optimal}
M.~P. Bends{\o}e, Optimal shape design as a material distribution problem,
  Structural optimization 1~(4) (1989) 193--202.

\bibitem{allaire2002level}
G.~Allaire, F.~Jouve, A.-M. Toader, A level-set method for shape optimization,
  Comptes {R}endus {M}athematique 334~(12) (2002) 1125--1130.

\bibitem{chen2007shape}
J.~Chen, V.~Shapiro, K.~Suresh, I.~Tsukanov, Shape optimization with
  topological changes and parametric control, International {J}ournal for
  {N}umerical {M}ethods in {E}ngineering 71~(3) (2007) 313--346.

\bibitem{eschenauer1994bubble}
H.~A. Eschenauer, V.~V. Kobelev, A.~Schumacher, Bubble method for topology and
  shape optimization of structures, Structural optimization 8~(1) (1994)
  42--51.

\bibitem{novotny2003topological}
A.~A. Novotny, R.~A. Feij{\'o}o, E.~Taroco, C.~Padra, Topological sensitivity
  analysis, Computer methods in applied mechanics and engineering 192~(7-8)
  (2003) 803--829.

\bibitem{bourdin2003design}
B.~Bourdin, A.~Chambolle, Design-dependent loads in topology optimization,
  {ESAIM}: {C}ontrol, {O}ptimisation and {C}alculus of {V}ariations 9 (2003)
  19--48.

\bibitem{sokolowski2009topological}
J.~Sokolowski, A.~Zochowski, Topological derivative in shape optimization,
  Encyclopedia of {O}ptimization (2009) 3908--3918.

\bibitem{borrvall2001large}
T.~Borrvall, J.~Petersson, Large-scale topology optimization in {3D} using
  parallel computing, Computer methods in applied mechanics and engineering
  190~(46-47) (2001) 6201--6229.

\bibitem{wu2015system}
J.~Wu, C.~Dick, R.~Westermann, A system for high-resolution topology
  optimization, {IEEE} transactions on visualization and computer graphics
  22~(3) (2015) 1195--1208.

\bibitem{andreassen2011efficient}
E.~Andreassen, A.~Clausen, M.~Schevenels, B.~S. Lazarov, O.~Sigmund, Efficient
  topology optimization in {MATLAB} using 88 lines of code, Structural and
  {M}ultidisciplinary {O}ptimization 43~(1) (2011) 1--16.

\bibitem{liu2014efficient}
K.~Liu, A.~Tovar, An efficient {3D} topology optimization code written in
  {M}atlab, Structural and {M}ultidisciplinary {O}ptimization 50~(6) (2014)
  1175--1196.

\bibitem{jang2008design}
I.~G. Jang, B.~M. Kwak, Design space optimization using design space adjustment
  and refinement, Structural and {M}ultidisciplinary {O}ptimization 35~(1)
  (2008) 41--54.

\bibitem{kim2012new}
S.~Y. Kim, I.~Y. Kim, C.~K. Mechefske, A new efficient convergence criterion
  for reducing computational expense in topology optimization: reducible design
  variable method, International {J}ournal {F}or {N}umerical {M}ethods {I}n
  {E}ngineering 90~(6) (2012) 752--783.

\bibitem{lynch2019machine}
M.~E. Lynch, S.~Sarkar, K.~Maute, Machine learning to aid tuning of numerical
  parameters in topology optimization, Journal of {M}echanical {D}esign
  141~(11) (2019).

\bibitem{sosnovika2017neural}
I.~Sosnovika, I.~Oseledetsb, {N}eural networks for topology optimization,
  arXiv:1709.09578 (2017).

\bibitem{lin2018investigation}
Q.~Lin, J.~Hong, Z.~Liu, B.~Li, J.~Wang, Investigation into the topology
  optimization for conductive heat transfer based on deep learning approach,
  International {C}ommunications in {H}eat and {M}ass {T}ransfer 97 (2018)
  103--109.

\bibitem{gaymann2019deep}
A.~Gaymann, F.~Montomoli, Deep neural network and monte carlo tree search
  applied to fluid-structure topology optimization, Scientific reports 9~(1)
  (2019) 1--16.

\bibitem{o2019standard}
N.~J. O'Neill, Standard and inception-based encoder-decoder neural networks for
  predicting the solution convergence of design optimization algorithms,
  Master's thesis, University of Colorado Boulder (2019).

\bibitem{zhang2019deep}
Y.~Zhang, A.~Chen, B.~Peng, X.~Zhou, D.~Wang, A deep {C}onvolutional {N}eural
  {N}etwork for topology optimization with strong generalization ability, arXiv
  preprint arXiv:1901.07761 (2019).

\bibitem{rawat2019application}
S.~Rawat, M.~H. Shen, Application of {A}dversarial {N}etworks for {3D}
  structural topology optimization, Tech. rep., {SAE} {T}echnical {P}aper
  (2019).

\bibitem{rawat2019novel}
S.~Rawat, M.-H.~H. Shen, A {N}ovel {T}opology {O}ptimization {A}pproach using
  {C}onditional {D}eep {L}earning, arXiv preprint arXiv:1901.04859 (2019).

\bibitem{shen2019new}
M.-H.~H. Shen, L.~Chen, A {N}ew {CGAN} {T}echnique for {C}onstrained {T}opology
  {D}esign {O}ptimization, arXiv preprint arXiv:1901.07675 (2019).

\bibitem{guo2018indirect}
T.~Guo, D.~J. Lohan, R.~Cang, M.~Y. Ren, J.~T. Allison, An indirect design
  representation for topology optimization using variational autoencoder and
  style transfer, in: 2018 {AIAA/ASCE/AHS/ASC} {S}tructures, {S}tructural
  {D}ynamics, and {M}aterials {C}onference, 2018, p. 0804.

\bibitem{lei2019machine}
X.~Lei, C.~Liu, Z.~Du, W.~Zhang, X.~Guo, Machine learning-driven real-time
  topology optimization under moving morphable component-based framework,
  Journal of {A}pplied {M}echanics 86~(1) (2019) 011004.

\bibitem{weiss2016survey}
K.~Weiss, T.~M. Khoshgoftaar, D.~Wang, A survey of transfer learning, Journal
  of {B}ig data 3~(1) (2016) 9.

\bibitem{torrey2010transfer}
L.~Torrey, J.~Shavlik, Transfer learning, in: Handbook of research on machine
  learning applications and trends: algorithms, methods, and techniques, IGI
  Global, 2010, pp. 242--264.

\bibitem{khatami2018sequential}
A.~Khatami, M.~Babaie, H.~R. Tizhoosh, A.~Khosravi, T.~Nguyen, S.~Nahavandi, A
  sequential search-space shrinking using cnn transfer learning and a radon
  projection pool for medical image retrieval, Expert {S}ystems with
  {A}pplications 100 (2018) 224--233.

\bibitem{hossain2018multiclass}
I.~Hossain, A.~Khosravi, I.~Hettiarachchi, S.~Nahavandi, Multiclass informative
  instance transfer learning framework for motor imagery-based brain-computer
  interface, Computational {I}ntelligence and {N}euroscience 2018 (2018).

\bibitem{Pan2010survey}
S.~J. Pan, Q.~Yang, W.~Fan, S.~J.~P. (ph. D), A survey on transfer learning,
  {IEEE} {T}ransactions on Knowledge and Data Engineering (2010).

\bibitem{le2010stress}
C.~Le, J.~Norato, T.~Bruns, C.~Ha, D.~Tortorelli, Stress-based topology
  optimization for continua, Structural and {M}ultidisciplinary {O}ptimization
  41~(4) (2010) 605--620.

\bibitem{sigmund2007morphology}
O.~Sigmund, Morphology-based black and white filters for topology optimization,
  Structural and {M}ultidisciplinary {O}ptimization 33~(4-5) (2007) 401--424.

\bibitem{Steiner2001}
G.~Steiner, Transfer of learning, cognitive psychology of, in: N.~J. Smelser,
  P.~B. Baltes (Eds.), International Encyclopedia of the Social \& Behavioral
  Sciences, Pergamon, 2001, pp. 15845 -- 15851.

\bibitem{mccann2017convolutional}
M.~T. McCann, K.~H. Jin, M.~Unser, Convolutional neural networks for inverse
  problems in imaging: A review, {IEEE} {S}ignal {P}rocessing {M}agazine 34~(6)
  (2017) 85--95.

\bibitem{lehman2010revising}
J.~Lehman, K.~O. Stanley, Revising the evolutionary computation abstraction:
  minimal criteria novelty search, in: Proceedings of the 12th annual
  conference on Genetic and evolutionary computation, ACM, 2010, pp. 103--110.

\bibitem{bourdin2001filters}
B.~Bourdin, Filters in topology optimization, International {J}ournal {F}or
  {N}umerical {M}ethods {I}n {E}ngineering 50~(9) (2001) 2143--2158.

\bibitem{christen2007quality}
P.~Christen, K.~Goiser, Quality and complexity measures for data linkage and
  deduplication, in: Quality measures in data mining, Springer, 2007, pp.
  127--151.

\bibitem{birch1983classroom}
J.~Birch, T.~Robertson, A classroom note on the sample variance and the second
  moment, The {A}merican {M}athematical {M}onthly 90~(10) (1983) 703--705.

\bibitem{cang2019one}
R.~Cang, H.~Yao, Y.~Ren, One-shot generation of near-optimal topology through
  theory-driven machine learning, Computer-Aided Design 109 (2019) 12--21.

\end{thebibliography}

\end{document}